\pdfoutput=1

\documentclass[11pt]{article}

\usepackage{acl}

\usepackage{newtxtext,newtxmath}
\usepackage{colortbl}
\usepackage{xcolor}
\usepackage{latexsym}
\usepackage{amsmath}
\usepackage{booktabs}
\usepackage{siunitx}
\usepackage{multirow}
\usepackage{array}
\usepackage{caption}
\usepackage{subfigure}
\usepackage{subcaption}
\usepackage{amsmath}
\usepackage{tikz}
\usepackage{pgfplots}
\usepackage{pgf-pie}
\pgfplotsset{compat=1.18}
\usepackage{amsfonts}
\usepackage{mdframed}
\usepackage{xcolor}
\usepackage{tcolorbox}
\usepackage{ifthen}
\usepackage{titlesec}
\usepackage{multicol} 
\usepackage{adjustbox}
\usepackage{tikz}
\usepackage{twemojis}
\usepackage{longtable}
\usepackage{graphicx}
\usepackage{longtable}
\usepackage{enumitem}

\newcommand{\hl}[1]{\textcolor{green!50!black}{#1}}
\newcommand{\hlred}[1]{\textcolor{red}{#1}}

\usepackage{float} 
\definecolor{llmwhite}{HTML}{FFFFFF} 
\definecolor{llmframe}{HTML}{AAAAAA} 
\definecolor{llmtitle}{HTML}{000000} 
\newtcolorbox{llmpromptstyle}[1][]{
  colback=llmwhite, 
  colframe=llmframe,
  coltitle=llmtitle,
  fonttitle=\bfseries\sffamily\small,
  title=#1,
  rounded corners,
  fontupper=\sffamily\small,
  before upper={\raggedright},
  boxrule=0.5pt,
  left=5pt, right=5pt, top=3pt, bottom=3pt,
  nobeforeafter
}
\tcbuselibrary{breakable}
\newenvironment{llmprompt}[1][]{
  \begin{figure*}[!h] 
  \centering
  \begin{llmpromptstyle}[#1] 
}{
  \end{llmpromptstyle}
  \end{figure*} 
}
\author{
Abhilekh Borah$^{1}$, Shubhra Ghosh$^{2,*}$, Kedar Joshi$^{1,*}$,
\textbf{Aditya Kumar Guru}$^{1,*}$,\\
\textbf{Kripabandhu Ghosh}$^{3}$ \\
$^{1}$Manipal University Jaipur,
$^{2}$Indian Institute of Technology, Patna, \\
$^{3}$Indian Institute of Science Education and Research, Kolkata \\
}

\title{\raisebox{-0.3em}{\twemoji[height=1.3em]{1f575}}\enspace Don't Judge a Book by its Cover: Testing LLMs' Robustness Under Logical Obfuscation}
\begin{document}
\maketitle
\renewcommand{\thefootnote}{\fnsymbol{footnote}}
\footnotetext[1]{Equal Contribution.}
\begin{abstract}
Tasks such as solving arithmetic equations, evaluating truth tables, and completing syllogisms are handled well by large language models (LLMs) in their standard form, but they often fail when the same problems are posed in logically equivalent yet obfuscated formats. To study this vulnerability, we introduce \textbf{Logifus\footnote[2]{https://github.com/abhilekhborah/logical-obfuscation}}, a structure-preserving logical obfuscation framework, and, utilizing this, we present \textbf{LogiQAte\footnote[3]{https://huggingface.co/datasets/abhilekhborah/LogiQAte}}, a first-of-its-kind diagnostic benchmark with 1{,}108 questions across four reasoning tasks: \text{(i) Obfus FOL} (first-order logic entailment under equivalence-preserving rewrites), \text{(ii) Obfus Blood Relation} (family-graph entailment under indirect relational chains), \text{(iii) Obfus Number Series} (pattern induction under symbolic substitutions), and \text{(iv) Obfus Direction Sense} (navigation reasoning under altered directions and reference frames). Across all the tasks, evaluating six state-of-the-art models, we find that obfuscation severely degrades zero-shot performance, with performance dropping on average by 47\% for GPT-4o, 27\% for GPT-5, and 22\% for reasoning model, o4-mini. Our findings reveal that current LLMs parse questions without deep understanding, highlighting the urgency of building models that genuinely comprehend and preserve meaning beyond surface form.
\end{abstract}
\begin{figure}[t]
\centering
\includegraphics[width=\columnwidth]{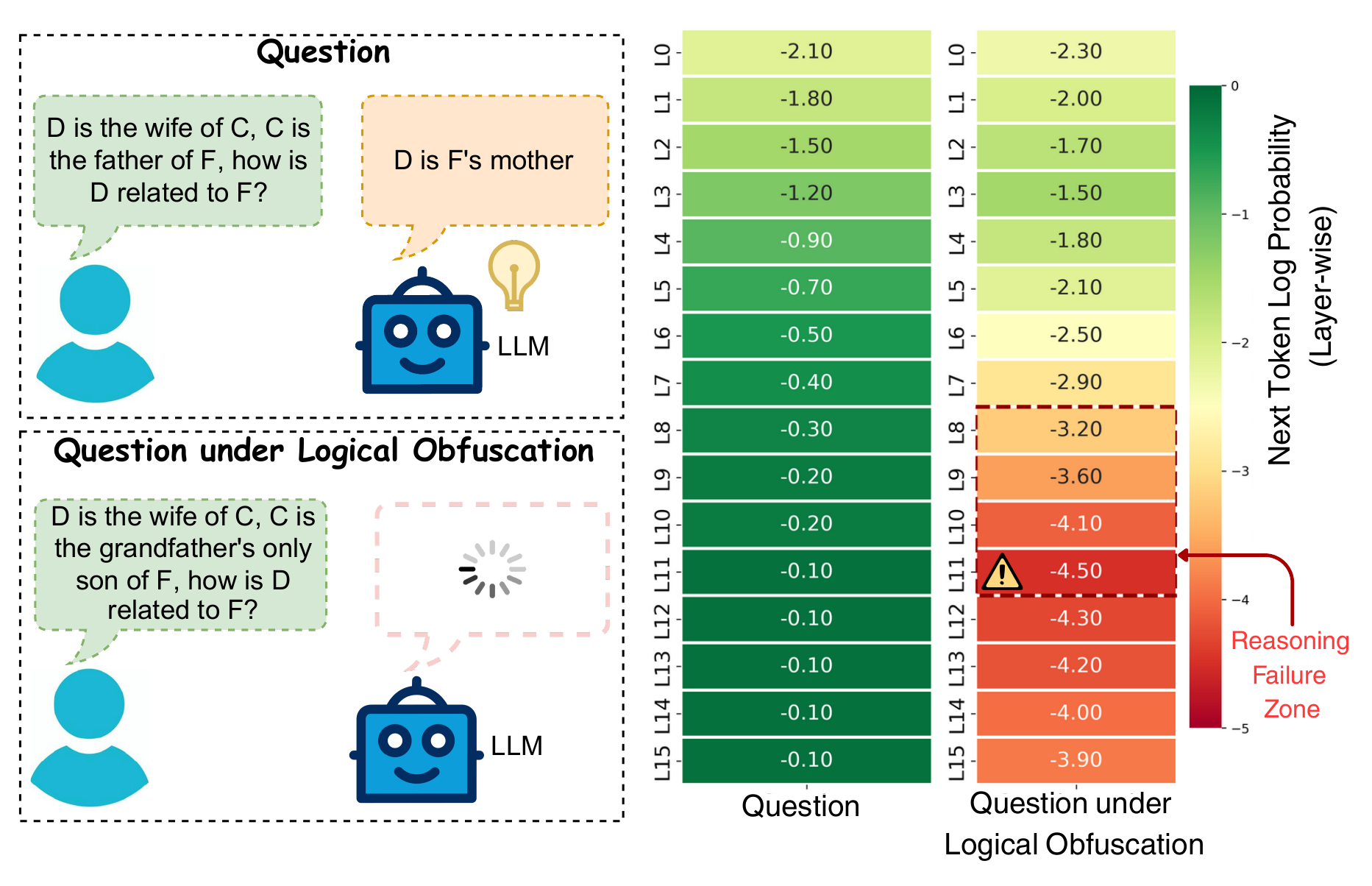}
\caption{\small \textbf{An Overview of LLMs under Logical Obfuscation with Logifus.} 
On the left, an obfuscated version of a question, while remaining \textit{logically equivalent}, poses significant challenge to the model. 
On the right, we track the model’s confidence in predicting the correct next word or token, quantified as the layer-wise next-token log probability assigned to that token at each step. Under logical obfuscation, these layer-wise scores drop sharply as the question is processed, indicating a transition into a reasoning-failure zone (cf. Section \ref{layerwise}). The plot shows this degradation across the first 15 layers of LLaMA 3.1 8B~\cite{touvron2023llama}.}
\label{intronewfig}
\vspace{-7mm}
\end{figure}
\vspace{-2mm}
\section{Introduction}
\vspace{-2mm}
\textit{``A large language model can’t do deductive reasoning. It’s set up to do pattern recognition and react to those patterns''}, warned Brent Smolinski, IBM's Vice President and Senior Partner for Global AI Strategy ~\cite{SmolinksiIBM2024}. Complementing this critique, in 2024, MIT researchers demonstrate that generative models “produce impressive outputs without recovering the underlying world model,” crashing when simple detours arise in a New York City navigation task~\cite{RambachanMIT2024}. Together, these observations underscore a critical question of whether large language models (LLMs) truly \textit{reason} or simply \textit{memorize} and match patterns learned from their training data.

Logical reasoning, the ability to derive new facts from known ones, solve puzzles, and adapt to novel situations, has anchored AI research since its inception. Early symbolic systems~\cite{McCarthy1959, eizenbaum1966, Winograd1971, Colmerauer1973, Shortliffe1976} pioneered rule-based, logic-driven approaches; more recently, chain-of-thought (CoT) prompting has enabled LLMs to articulate intermediate steps, improving performance on arithmetic, commonsense, and symbolic reasoning tasks~\cite{Wei2022ChainOfThought}. Yet, the question of genuine reasoning persists. State-of-the-art systems such as GPT-5~\cite{openai2025gpt5}, Claude~4~\cite{anthropic2025claude4}, and o4~\cite{openai2025o4} demonstrate impressive capabilities, but it remains unclear whether they construct and verify logical chains or merely exploit memorized patterns. Standard benchmarks (MMLU~\cite{hendrycks2020mmlu}, BIG-Bench Hard~\cite{srivastava2022beyond}, ARC~\cite{clark2018arc}) typically score only the final answer, allowing high performance via surface recall. Empirically, models often falter when equivalent questions are rephrased or obfuscated, underscoring the gap between pattern matching and robust reasoning~\cite{Xie2025Memorization}.

To understand reasoning beyond memorization and to check if the model is truly understanding the problem question, we explore \emph{logical obfuscation}, i.e., rewriting a problem into a different surface form while preserving its logical structure. As shown in Fig.~\ref{intronewfig}, in a simple family-relation puzzle, the sentence \emph{``D is the wife of C, C is the father of F''} can be obfuscated as \emph{``D is the wife of C, C is the grandfather's only son of F''}. Both describe the same relationship, yet an LLM that answers ``D is F's mother'' in the original question struggles when presented with the obfuscated version. Building on this idea, we present \textbf{LogiQAte}, the first-of-its-kind benchmark of 1{,}108 questions, built using \textbf{Logifus}, a proposed, structure-preserving obfuscation algorithm that systematically generates logically equivalent but syntactically diverse versions of each question. LogiQAte spans four tasks: \textbf{(i) Obfus FOL}, which evaluates first-order logic entailment under equivalence-preserving rewrites (e.g., replacing ``If it rains, the ground gets wet'' with ``The ground does not stay dry whenever it rains''), \textbf{(ii) Obfus Blood Relation}, which tests family-graph reasoning under obfuscation with indirect kinship chains and alternative terms (cf. Fig \ref{intronewfig}), \textbf{(iii) Obfus Number Series}, which requires inductive pattern detection in numerical sequences obfuscated via symbolic substitutions and arithmetic transformations (e.g., the base sequence $2,4,6,8,\ ?$ rewritten as $n, n+2, n+4, n+6,\ ?$), and \textbf{(iv) Obfus Direction Sense}, where navigation puzzles are rephrased with altered directions and frames of reference (e.g., replacing ``turn left and walk 3 steps'' with ``move westward three paces''). 

\begin{figure*}[htbp]
    \centering
    \includegraphics[width=\textwidth]{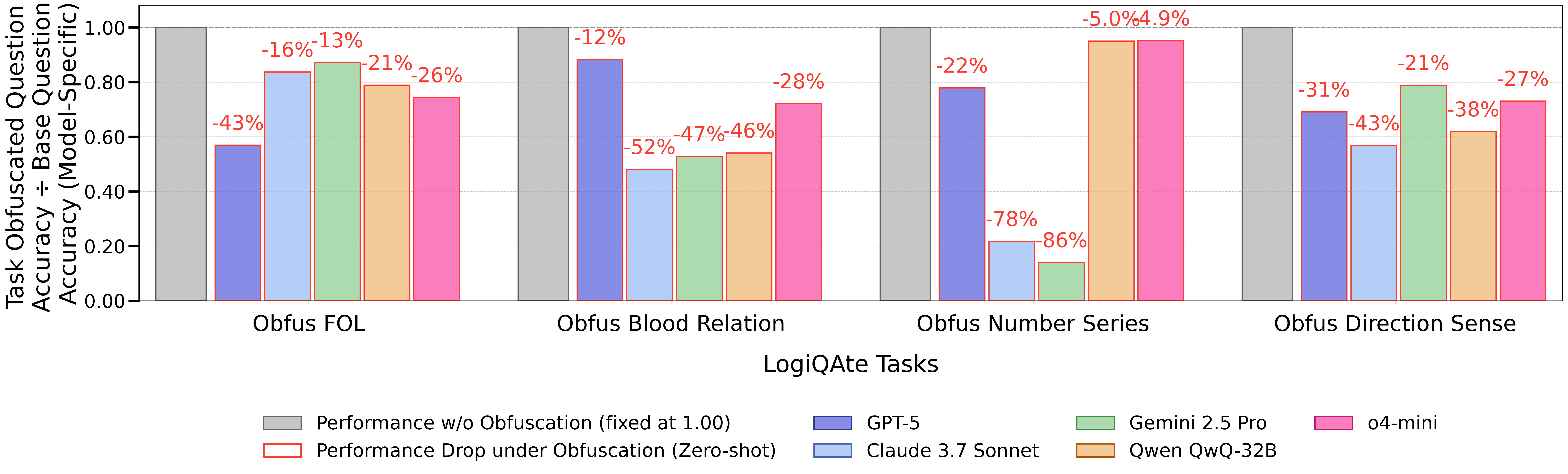}
    \vspace{-2mm}
    \caption{
        \small \textbf{Logical obfuscation degrades performance across tasks in LogiQAte.} 
        We report relative accuracy for obfuscated questions against base questions 
        (performance fixed at 1.00) across five LLMs under \textit{zero-shot} settings. Across all tasks and settings, we 
        observe consistent degradation, up to $-86\%$, highlighting the brittleness 
        of LLM reasoning under logical obfuscation.
    }
    \label{fig:accuracy_comparison}
    \vspace{-5mm}
\end{figure*}

Across six state-of-the-art models, we observe a consistent average performance drop of 28.8\% under logical obfuscation, computed across all tasks and averaged over zero-shot, few-shot, and CoT settings (cf. Fig.~\ref{fig:accuracy_comparison}). To understand this brittleness, we examine \emph{why} performance degrades. Memorization tests show detection rates rising from 50\% on original to 82\% on obfuscated questions, indicating heavy reliance on memorization rather than reasoning. Layer-wise next-token log-probability analysis further reveals a 50–80\% collapse in token-level confidence in late layers, critical for reasoning, relative to base questions (cf. Section~\ref{layerwise}).
The paper makes the following contributions:
\vspace{-2mm}
\begin{enumerate}
    \item We introduce \textbf{Logifus}, a novel structure-preserving algorithm for generating logically obfuscated questions across four reasoning tasks.  
    \vspace{-2mm}
    \item We present \textbf{LogiQAte}, a first-of-its-kind diagnostic benchmark comprising 1{,}108 obfuscated reasoning questions spanning four reasoning tasks (cf. Fig. \ref{fig:mainfig}), conducting a comprehensive empirical analysis across nine open- and closed-source (general purpose and reasoning-focused) LLMs. 
    \vspace{-4mm}
    \end{enumerate}



\vspace{-1mm}
\section{Related Works}
Research on evaluating the reasoning capabilities of LLMs spans several paradigms. Natural language inference (NLI) benchmarks such as LogicNLI generate large-scale synthetic triplets \cite{tian2021diagnosing}, while FOLIO provides 1,430 human-written examples annotated with first-order logic \cite{han2022folio}. More complex deductive datasets, including LogicBench \cite{parmar2024logicbench}, ProofWriter \cite{tafjord2021proofwriter}, Multi-LogiEval \cite{patel2024multi}, and ProntoQA \cite{madaan2022language}, explore varied inference types but typically evaluate on clean formulations. Machine reading comprehension datasets such as LogiQA \cite{liu2020logiqa}, ReClor \cite{yu2020reclor}, and LingOly \cite{bean2024lingoly} similarly target reasoning in QA or linguistic puzzles without testing robustness to equivalent reformulations.

\paragraph{Obfuscation Studies.}
ObfusQAte proposes multi-tier semantic transformations for factual QA and reports frequent hallucinations and failures under these variants~\cite{ghosh2025obfusqate}. In code generation, MetamorphASM evaluates resilience to obfuscated assembly and source, finding that large code models can reduce superficial complexity yet often break semantics under heavy obfuscation~\cite{mohseni2025metamorphasm}. 
On the other hand, existing benchmarks such as GSM-Symbolic~\cite{mirzadeh2025gsm_symbolic} target a different failure mode (i.e., parameter variation under a fixed reasoning template). In this setting, the logical form is kept unchanged (the reasoning structure remains the same), and only surface entities or values are replaced, thus testing generalization across different instantiations of the same template. In contrast, LogiQAte varies the logical structure itself (i.e., how relations are expressed and composed) while keeping the truth conditions invariant. 

Together, these works show that obfuscation is a stringent robustness test, but prior efforts center mainly on factual QA, code or parameter variation under fixed reasoning templates, leaving robustness to logical obfuscation under equivalence largely unexplored. \text{LogiQAte} addresses this gap by introducing the first benchmark to apply logical obfuscation via equivalence-preserving rewrites to assess reasoning robustness across deductive, relational, numerical, directional, and spatial tasks.

\section{Proposed Setup}\label{ProposedSetup}
To rigorously evaluate the robustness of LLMs' logical reasoning, we introduce a novel prompting framework called \textbf{Logifus}. The central principle of Logifus is the application of controlled, logic-preserving obfuscation. This involves transforming questions and premises to alter their surface-level linguistic or structural representation while keeping the underlying logical structure and, consequently, the correct answer invariant. We define an obfuscation function $f$ as a mapping $f: Q \rightarrow Q'$, where $Q = \{q_1, q_2, \dots, q_n\}$ denotes a set of questions and $Q'$ is its {\it logically equivalent} obfuscated counterpart. For any question $q_i \in Q$ and its corresponding {\it logically equivalent} obfuscated version $q'_i \in Q'$, both share the same correct solution as they are logically equivalent. In other words, if $q_i$ has a valid solution $M_i$, then $q'_i$ yields the same solution $M_i$, even though their question's surface forms differ because $q'_i$ is {\it logically equivalent} to $q'_i$, compelling models to engage in deeper, more authentic reasoning processes in understanding questions. We report all \text{Logifus} algorithms in Appendix~\ref{logifusalgos}.

\begin{figure*}[htbp]
    \centering
    \includegraphics[width=\linewidth]{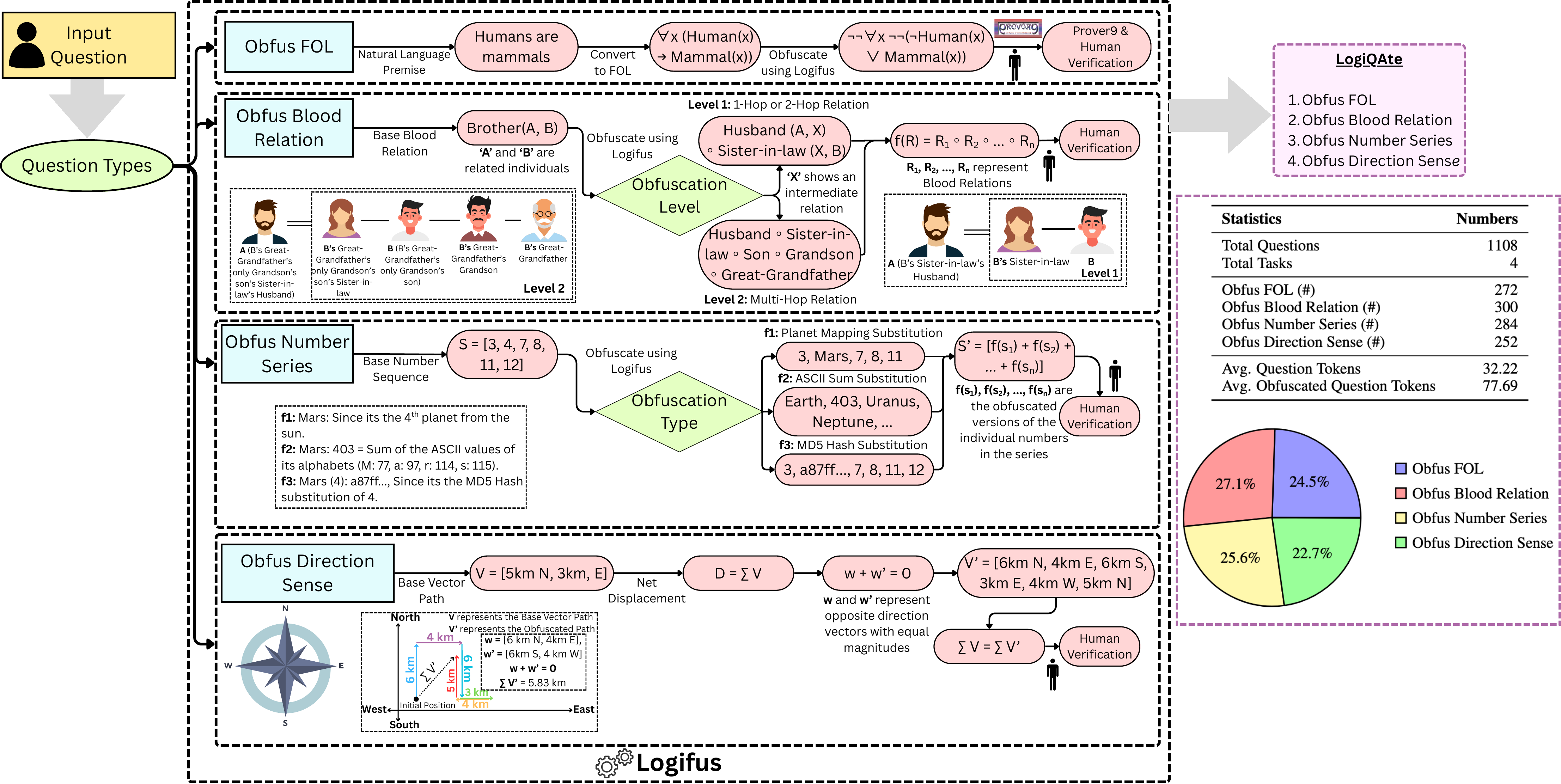}
\caption{\small \textbf{Overview of \text{LogiQAte} and Logifus.} 
From a user question (left), four task types are generated: Obfus FOL, Obfus Blood Relation, Obfus Number Series, and Obfus Direction Sense. Each task applies a logical equivalence-preserving transformation and undergoes human (along with Prover9 \cite{prover9-mace4} in the case of Obfus FOL) verification. The right panel summarizes the dataset statistics: 1{,}108 questions across four tasks.}

    \label{fig:mainfig}
    \vspace{-5mm}
\end{figure*}

Leveraging the Logifus framework, we introduce \textbf{LogiQAte}, a novel diagnostic benchmark comprising 1,108 questions, designed and instantiated across four complementary reasoning tasks to quantitatively assess a model's sensitivity to logical obfuscations. To guarantee the benchmark's integrity, every base question and its corresponding obfuscated version underwent a stringent validation process involving a two-human-annotator loop. This ensured that the logical equivalence and the original conclusion were preserved, achieving near-perfect agreement across all tasks (Cohen’s $\kappa > 0.88$). The task types are visualized in Fig. \ref{fig:mainfig} and described below.
\subsection{Obfus FOL}\label{ObfusFOL}
\noindent\textbf{Task Description.} First-order logic (FOL) serves as a gold standard for evaluating structured, formal reasoning. Our Obfus First-Order Logic task, comprising 272 questions, applies the Logifus framework to FOL entailment problems to probe a model's ability to handle logically complex rewrites. This is achieved through \textit{FOL Transformation}, where we generate obfuscated variants by rephrasing premises using equivalent logical formulations, such as applying De Morgan's laws or double-negation elimination. For instance, a premise like \textit{“Humans are mammals”} ($\forall x\,(\mathit{Human}(x) \to \mathit{Mammal}(x))$) is transformed into a convoluted but logically equivalent form like \textit{“It is false that if there exists an entity that is not an animal, then there exists an entity that is a human who is not a mammal”}.

\noindent\textbf{Dataset Construction.} The construction of questions for the Obfus FOL task started with entries from the FOLIO \cite{han2022folio} dataset; we first used GPT-4o \cite{openai2024gpt4o} to simplify the natural language premises, creating a clearer, less verbose baseline. To safeguard against potential AI-generated inaccuracies, these simplified premises were cross-verified by two human annotators, achieving a strong inter-annotator agreement (Cohen's $\kappa=0.95$). For a higher degree of certainty, we employed Prover9 \cite{prover9-mace4}, an automated theorem prover, to provide a formal guarantee of logical equivalence (cf. Appendix for procedure). Only after this rigorous baseline validation were the premises obfuscated, and a final manual verification by two annotators confirmed the logical integrity of the obfuscated versions, yielding a high agreement score (Cohen's $\kappa=0.89$).
\begin{mdframed}[
    linecolor=black,
    linewidth=0.5pt,
    topline=true, bottomline=true,
    leftline=false, rightline=false,
    innerleftmargin=0pt, innerrightmargin=0pt,
    innertopmargin=2pt, innerbottommargin=2pt,
]
\textbf{FOL Transformation} \\
$\blacktriangleright$ \textbf{Input}: A set of premises $\mathcal{P}$ and a conclusion $\mathcal{C}$ where $\mathcal{P} \models \mathcal{C}$. \\
$\blacktriangleright$ \textbf{Transformation}: Generate an obfuscated premise set $\mathcal{P}'$ using logical equivalences (e.g., De Morgan's laws) such that $\mathcal{P} \equiv \mathcal{P}'$. \\
$\blacktriangleright$ \textbf{Output}: An obfuscated problem $(\mathcal{P}', \mathcal{C})$ where the entailment $\mathcal{P}' \models \mathcal{C}$ still holds.
\end{mdframed}
\vspace{-2mm}
\subsection{Obfus Blood Relation}\label{Blood relation obfuscation}
\noindent\textbf{Task Description.} Blood-relation puzzles are a classic benchmark for relational reasoning, which involves understanding and reasoning about relationships among entities, people, or objects. Our Obfus Blood Relation task, consisting of 300 questions, evaluates these skills using a technique we term \textit{Relational Obfuscation}. This process involves systematically transforming straightforward kinship relations into indirect, multi-hop descriptions that preserve the underlying family structure, thereby increasing linguistic and cognitive complexity. For example, the direct relation \textit{“A is the brother of B”} is transformed in \textbf{Obfuscation Level 1} to a two-step relation like \textit{“A is the sister-in-law’s husband of B.”} In \textbf{Obfuscation Level 2}, the complexity is further increased with a longer reasoning chain, such as \textit{“A is the great-grandfather’s only grandson’s son’s sister-in-law’s husband of B.”} These levels systematically increase the cognitive load, allowing us to measure a model's breaking point.

\noindent\textbf{Dataset Construction.} The construction of questions for the Obfus Blood Relation task, was done by manually curating puzzles from reputable educational platforms such as GeeksforGeeks, IndiaBix and Brilliant. For each puzzle, one key relational premise was transformed using our Level 1 and Level 2 obfuscation strategies. The logical equivalence of every generated obfuscation was manually confirmed by two human annotators, who cross-checked the obfuscation type and logical integrity, achieving a strong inter-annotator agreement score (Cohen's $\kappa$ = 90\%).

\begin{mdframed}[
    linecolor=black,
    linewidth=0.5pt,
    topline=true, bottomline=true,
    leftline=false, rightline=false,
    innerleftmargin=0pt, innerrightmargin=0pt,
    innertopmargin=2pt, innerbottommargin=2pt,
]
\textbf{Relational Obfuscation} \\
$\blacktriangleright$ \textbf{Input}: A direct kinship statement (e.g., “A is the brother of B”). \\
$\blacktriangleright$ \textbf{Transformation}: Rewrite the statement as an indirect, multi-hop description (e.g., “A is the sister-in-law’s husband of B”). \\
$\blacktriangleright$ \textbf{Output}: An obfuscated puzzle with increased reasoning chain length but an unchanged underlying family graph.
\end{mdframed}
\vspace{-2mm}
\subsection{Obfus Number Series}\label{numberseries}
\noindent\textbf{Task Description.} Our task Obfus Number Series, containing \text{284} questions, evaluates abstract pattern recognition and inductive reasoning, by applying Logifus through a technique, we term as \textit{Pattern-Preserving Obfuscation}, which is defined as disguising the surface representation of numbers without altering the underlying mathematical pattern of the sequence. This type is organized into 3 distinct parts, each derived from the same set of base sequences but transformed using different obfuscation strategies, detailed below:

\textbf{(i) Type 1: Planet-Name Substitution.}  
We replace each integer in the sequence with a fixed, human-readable token drawn from a small vocabulary of celestial body names (e.g., \emph{Mercury}, \emph{Venus}, \emph{Earth}, \emph{Mars}, \ldots). The mapping is one-to-one and deterministic: for example, the integer \texttt{4} is always rendered as \emph{Mars}. This substitution preserves ordinal relationships but hides the numeric form, forcing the model to treat tokens as abstract symbols that nevertheless encode an underlying numeric mapping.
\textit{Example:} the base sequence \texttt{[2, 4, 6, 8]} becomes \{\emph{Venus}, \emph{Mars}, \emph{Jupiter}, \emph{Saturn}\} under a fixed mapping (2$\to$Venus, 4$\to$Mars, 6$\to$Jupiter, 8$\to$Saturn).
    
\textbf{(ii) Type 2: ASCII Sum Substitution.}  
After performing the planet-name substitution (Type 1), we optionally convert each planet label to a numeric encoding equal to the sum of its characters' ASCII (or Unicode codepoint) values. For instance, the string ``Mars'' becomes the integer 403 if the ASCII sum of `M' + `a' + `r' + `s' equals 403. This produces a purely numeric sequence that is semantically detached from the original numbers (the numeric values no longer follow the original progression), but the mapping from original number $\to$ planet name $\to$ ASCII-sum is deterministic and reversible during dataset construction. This transformation adds a layer of symbol-to-number encoding that is nontrivial for models to invert without recognizing the intermediate symbol mapping.
\textit{Example:} \emph{Mars} $\to$ ASCII sum 403, \emph{Venus} $\to$ ASCII sum 541, so the visible sequence could read \texttt{[541, 403, 857, ...]} even though the latent pattern was an arithmetic progression.
    
\textbf{(iii) Type 3: MD5 Hash Substitution.}  
For the strongest surface obfuscation, we replace each original number (or its planet-name token) with its 32-character MD5 hash string (hexadecimal). The MD5 mapping is deterministic but cryptographic in appearance (e.g., the digit `4' might map to \texttt{a87ff6...}). Because MD5 outputs look like random character sequences, this substitution makes the visible sequence appear entirely unstructured unless the model (or an auxiliary process) can invert or otherwise reason about the hash mapping. We use MD5 to simulate opaque, high-entropy encodings that a model must either ignore or decode via an external mapping.
\textit{Example:} base element \texttt{4} $\to$ planet token \emph{Mars} $\to$ MD5 hash \texttt{a87ff6...}; the sequence shown to the model becomes \{\texttt{d41d8c...}, \texttt{a87ff6...}, \texttt{0cc175...}, \dots\}.
\begin{mdframed}[
  linecolor=black,
  linewidth=0.5pt,
  topline=true, bottomline=true,
  leftline=false, rightline=false,
  leftmargin=0pt, rightmargin=0pt,
  innerleftmargin=0pt, innerrightmargin=0pt,
  innertopmargin=2pt, innerbottommargin=2pt
]
\begin{minipage}{\linewidth}
\textbf{Pattern-Preserving Obfuscation} \\
$\blacktriangleright$ \textbf{Input:} A numerical sequence $S = \{n_1, n_2, \dots, n_k\}$ (e.g., an arithmetic or geometric sequence). \\
$\blacktriangleright$ \textbf{Transformation:} Apply a deterministic mapping pipeline to each element: (i) map $n_i$ to a planet-name token via a fixed lookup table (Type 1); (ii) optionally convert that token to its ASCII-sum integer (Type 2); (iii) optionally replace the token or ASCII-sum with its MD5 hex string (Type 3). The pipeline may use one, two, or all three steps to produce progressively stronger obfuscation. \\
$\blacktriangleright$ \textbf{Output:} An obfuscated sequence $S' = \{f(n_1), f(n_2), \dots, f(n_k)\}$ that is visibly different from $S$ but preserves a reversible relationship to $S$ for dataset generation and verification.
\end{minipage}
\end{mdframed}

\noindent\textbf{Dataset Construction.} The \text{Obfus Number Series} dataset was built from standard aptitude and reasoning sources (e.g., IndiaBix, Brilliant.org, test-prep materials). A hardcoded python script (a) ingests canonical sequences, (b) applies predefined mappings to generate obfuscated variants, and (c) logs forward and reverse mappings for exact verification and answer reconstruction. Each obfuscated instance was manually reviewed to ensure the original generative rule remained uniquely recoverable under the allowed decoding procedure.

\subsection{Obfus Direction Sense}\label{ObfusDirection}
\noindent\textbf{Task Description.} Directional sense puzzles are fundamental for assessing spatial reasoning, such as path tracing, orientation tracking, left–right turn reasoning, and displacement calculation. Our Obfus Direction Sense task, consisting of 252 questions, utilizes a method we call \textit{Spatial Obfuscation}. This technique introduces complexity by embedding irrelevant detours and superfluous conditions within a set of directions, while ensuring the final destination remains identical to the original problem. For instance, a simple path such as \textit{“A person walks 5 km north and then 3 km east”} can be transformed into a longer, more complex version: \textit{“A person walks 6 km north, then 4 km east, 6 km south, 3 km east, 4 km west, and finally 5 km north”}. The additional movements, such as walking 6 km north and later 6 km south, are self-canceling detours that make the route appear more intricate while keeping the final position unchanged.
\begin{mdframed}[
    linecolor=black,
    linewidth=0.5pt,
    topline=true, bottomline=true,
    leftline=false, rightline=false,
    innerleftmargin=0pt, innerrightmargin=0pt,
    innertopmargin=2pt, innerbottommargin=2pt,
]
\textbf{Spatial Obfuscation} \\
$\blacktriangleright$ \textbf{Input}: A sequence of movement vectors $V = \{v_1, v_2, \dots, v_n\}$. \\
$\blacktriangleright$ \textbf{Transformation}: Insert self-canceling vector pairs $(w, w^{-1})$ into the sequence, where $w + w^{-1} = 0$. \\
$\blacktriangleright$ \textbf{Output}: An obfuscated sequence $V' = \{v_1, w, \dots, w^{-1}, v_n\}$ where the net displacement is unchanged: $\sum V = \sum V'$.
\end{mdframed}
\noindent\textbf{Dataset Construction.} For Obfus Direction Sense, we collected base directional puzzles from widely used logic and reasoning sources (e.g., IndiaBix, Brilliant, and general aptitude sites). For each base puzzle, a simple Python program generated obfuscated variants by randomly inserting paired detours (e.g., “go X east, then X west,” or “go Y north, then Y south”) at selected positions, directions, or magnitudes in the walking sequence. These pairs cancel out, preserving the final position and orientation; magnitudes and insertion points were varied to create multiple obfuscations per base. Each base-obfuscated pair was manually reviewed by two human annotators to verify unchanged displacement and absence of ambiguity, yielding strong inter-annotator agreement (Cohen’s $\kappa = 0.88$).
\section{Evaluation}

\subsection{Setup}
We evaluate nine state-of-the-art models (both open- and closed-source) on \text{LogiQAte}, grouped as follows: \text{general-purpose LLMs}: GPT-4o~\cite{openai2024gpt4o} and Claude~3.7~Sonnet~\cite{AnthropicClaude37Sonnet}; \text{reasoning-focused LLMs}: GPT-5~\cite{openai2025gpt5}, o4-mini~\cite{openai2025o4mini}, Gemini~2.5~Pro~\cite{gemini2025v25report}, and Qwen~QwQ-32B~\cite{qwen2025qwq32b}. We also evaluate \text{lower-parameter models}: GPT-4o-mini~\cite{openai2025o4mini}, Gemma-3-27B-IT~\cite{google2025gemma3}, and Llama-4-Maverick-17B~\cite{touvron2025llama4maverick} on our dataset (cf.\ Appendix~\ref{smallmodels}). All models are evaluated across zero-shot, few-shot, and chain-of-thought (CoT) settings with a temperature of 0.0.

\paragraph{Evaluation Metric.} Model performance is measured using Exact Match (EM) Accuracy, which computes the percentage of samples where the model’s normalized prediction, $\hat{y}_i$, exactly matches the normalized ground truth, ${y}_i$: for a task of $N$ evaluation samples, $\mathrm{EM}$ Accuracy = $\frac{1}{N}\sum_{i=1}^{N}\mathbb{1}\{\mathrm{norm}(\hat{y}_i)=\mathrm{norm}(y_i)\}$. Normalization involves standardizing case, removing punctuation, and ensuring whitespace uniformity to make comparisons resilient to minor textual variations (e.g., \text{``Sister-in-law''} $\equiv$ \text{``sister in law''}).
\begin{table*}[t]
\centering
\footnotesize
\setlength{\tabcolsep}{4.2pt}
\renewcommand{\arraystretch}{1.12}
\begin{adjustbox}{max width=\textwidth}
\begin{tabular}{l*{4}{ccc}}
\toprule
\textbf{Task \& Question Variant} &
\multicolumn{3}{c}{\textbf{GPT-5}} &
\multicolumn{3}{c}{\textbf{o4-mini}} &
\multicolumn{3}{c}{\textbf{Qwen QwQ-32B}} &
\multicolumn{3}{c}{\textbf{Claude 3.7 Sonnet}} \\
\cmidrule(lr){2-4}\cmidrule(lr){5-7}\cmidrule(lr){8-10}\cmidrule(lr){11-13}
 & \textbf{Zero-shot} & \textbf{Few-shot} & \textbf{CoT}
 & \textbf{Zero-shot} & \textbf{Few-shot} & \textbf{CoT}
 & \textbf{Zero-shot} & \textbf{Few-shot} & \textbf{CoT}
 & \textbf{Zero-shot} & \textbf{Few-shot} & \textbf{CoT} \\
 & \multicolumn{9}{c}{\textit{(Reasoning Models)}} & \multicolumn{3}{c}{\textit{(No Reasoning)}} \\
\midrule
\multicolumn{13}{l}{\textit{Obfus FOL}} \\
Base        & 0.98 & 0.87 & \cellcolor{red!20}0.82 & 0.92 & 0.92 & 0.92 & \cellcolor{red!20}0.77 & \cellcolor{red!20}0.76 & 0.84 & 0.82 & 0.85 & \cellcolor{red!20}0.82 \\
Obfuscation & \cellcolor{red!20}0.56 & \cellcolor{red!20}0.59 & \cellcolor{red!20}0.61 & 0.68 & 0.63 & 0.67 & 0.61 & 0.63 & 0.66 & 0.69 & 0.72 & 0.68 \\
\midrule
\multicolumn{13}{l}{\textit{Obfus Blood Relation}} \\
Base        & 0.52 & 0.73 & 0.72 & 0.69 & 0.68 & 0.74 & \cellcolor{red!20}0.42 & \cellcolor{red!20}0.43 & \cellcolor{red!20}0.48 & 0.49 & 0.45 & 0.59 \\
Obfuscation L1    & 0.46 & 0.68 & 0.67 & 0.59 & 0.61 & 0.60 & \cellcolor{red!20}0.26 & \cellcolor{red!20}0.30 & \cellcolor{red!20}0.29 & \cellcolor{red!20}0.26 & 0.36 & \cellcolor{red!20}0.29 \\
Obfuscation L2    & 0.45 & 0.59 & 0.55 & 0.41 & 0.44 & 0.45 & \cellcolor{red!20}0.19 & \cellcolor{red!20}0.25 & \cellcolor{red!20}0.24 & 0.21 & 0.26 & \cellcolor{red!20}0.24 \\
\midrule
\multicolumn{13}{l}{\textit{Obfus Number Series}} \\
Base        & 0.77 & 0.64 & 0.65 & 0.60 & 0.64 & 0.64 & 0.55 & 0.59 & 0.59 & \cellcolor{red!20}0.28 & \cellcolor{red!20}0.35 & \cellcolor{red!20}0.36 \\
Obfuscation & 0.60 & 0.57 & 0.57 & 0.57 & 0.57 & 0.56 & 0.53 & 0.53 & 0.51 & \cellcolor{red!20}0.06 & \cellcolor{red!20}0.08 & \cellcolor{red!20}0.07 \\
\midrule
\multicolumn{13}{l}{\textit{Obfus Direction Sense}} \\
Base        & 0.64 & 0.72 & 0.63 & 0.62 & 0.69 & 0.60 & 0.59 & 0.65 & 0.56 & \cellcolor{red!20}0.29 & \cellcolor{red!20}0.40 & \cellcolor{red!20}0.36 \\
Obfuscation & 0.44 & 0.55 & 0.50 & 0.45 & 0.51 & 0.48 & 0.37 & 0.46 & 0.36 & \cellcolor{red!20}0.17 & \cellcolor{red!20}0.22 & \cellcolor{red!20}0.21 \\
\midrule
\textbf{Average Performance Degradation (\%)} &
\textcolor{red}{-27.00\%} & \textcolor{red}{-20.20\%} & \textcolor{red}{-18.70\%} &
\textcolor{red}{-21.40\%} & \textcolor{red}{-22.70\%} & \textcolor{red}{-22.00\%} &
\textcolor{red}{-27.50\%} & \textcolor{red}{-23.30\%} & \textcolor{red}{-26.10\%} &
\textcolor{red}{-47.50\%} & \textcolor{red}{-41.60\%} & \textcolor{red}{-48.70\%} \\
\bottomrule
\end{tabular}
\end{adjustbox}
\caption{\small Performance degradation is observed across four logical obfuscation tasks in LogiQAte, with average drops ranging from \textcolor{red}{18.70\%} to \textcolor{red}{48.70\%} compared to base question performance. GPT-5, o4-mini, and Qwen QwQ-32B operate under reasoning configuration, while Claude 3.7 Sonnet operates with no reasoning. Each model is evaluated under Zero-shot, Few-shot, and Chain-of-Thought (CoT) prompt settings. \colorbox{red!20}{Red} cells indicate the lowest score for each prompting strategy within each task row.}
\label{tab:obfuscation_performance_compact}
\vspace{-5mm}
\end{table*}
\vspace{-2mm}
\section{Results and Discussion}
\label{sec:results}
\begin{figure}[t] 
  \centering
  \includegraphics[width=0.8\linewidth]{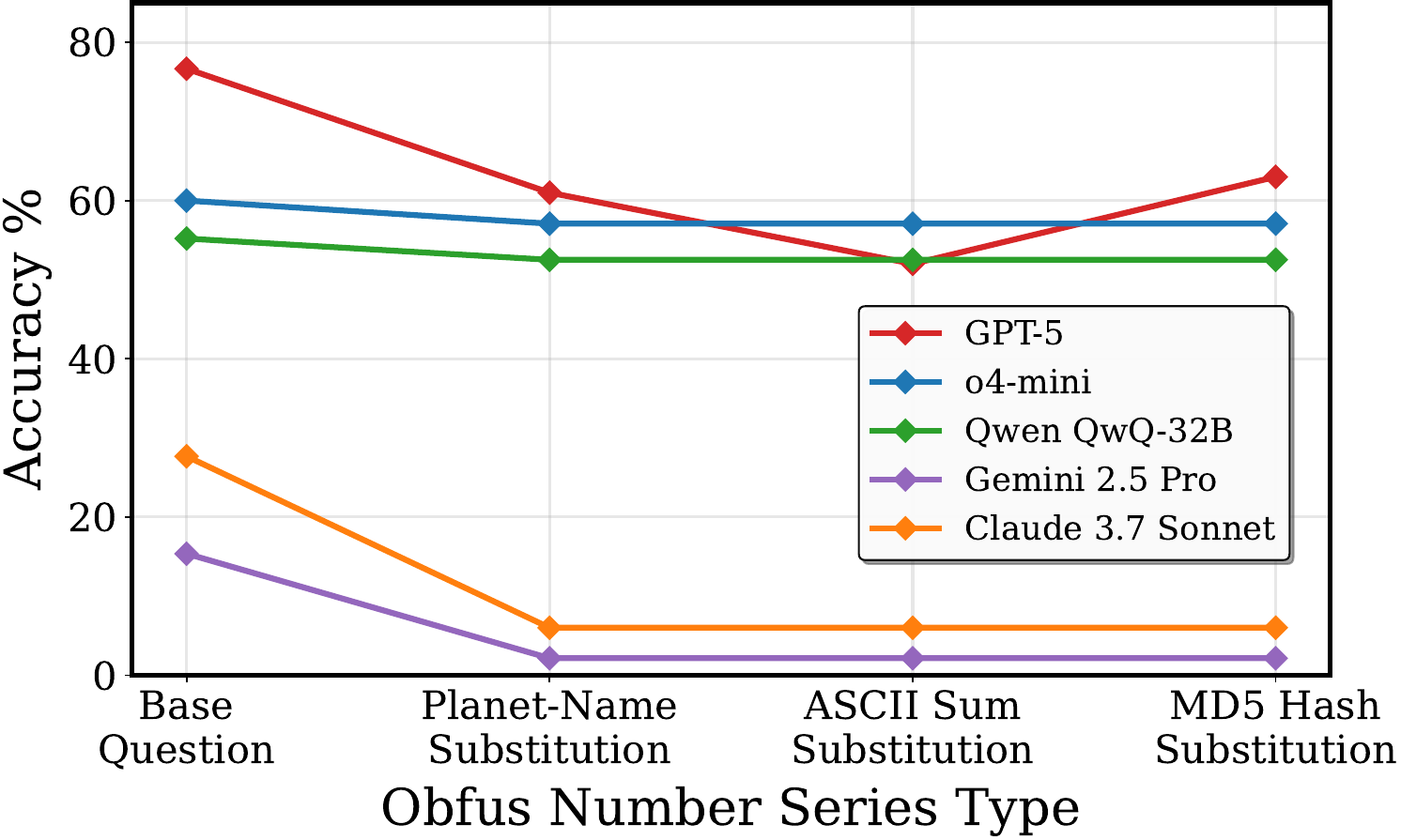} %
\vspace{-2mm}
\caption{
\textbf{\small Obfuscation reduces performance across types in Obfus Number Series.} Accuracy drops range from \textcolor{red}{20-45\%} for reasoning-focused models and up to \textcolor{red}{80-90\%} for general-purpose models such as Claude~3.7~Sonnet and Gemini~2.5~Pro, which already struggle on base questions. GPT-5 maintains higher accuracy across all obfuscation and base variants.
}

  \label{typenumberseries}
  \vspace{-7mm}
\end{figure}

Table~\ref{tab:obfuscation_performance_compact} and Table~\ref{tab:more_models_comparison} (cf. Appendix~\ref{moremodels}) summarize the performance of large reasoning and general-purpose models, while Table~\ref{tab:smaller_models_comparison} (cf. Appendix~\ref{smallmodels}) reports results for lower-parameter models, across all LogiQAte tasks under different prompting settings, and Appendix \ref{qualitativeanalysis} reports a qualitative analysis for each LogiQAte task. We provide a breakdown of zero-shot performance on the Obfus Number Series by obfuscation type in Fig.~\ref{typenumberseries}.
We set the QA level on par with undergraduate (UG) standards, validated by UG students who achieved near-perfect performance, up to 99\% accuracy on average across two human evaluators, on both base and obfuscated questions, while even highly capable reasoning models like GPT-5 struggled. This stark contrast casts a shadow on the reasoning ability of current LLMs. 

\noindent\textbf{Logical Obfuscation Reduces Performance Across Tasks, Prompt Settings, and Models.} In Table~\ref{tab:obfuscation_performance_compact}, average accuracy drops range from \text{18.70\%} to \text{48.70\%}, with Table~\ref{tab:more_models_comparison} showing a similar decline of \text{40.54\%}-\text{49.37\%}. Strikingly, on Obfus FOL, the reasoning-focused GPT-5 falls from 98\% to 56\% (zero-shot); on Obfus Number Series, GPT-4o drops by 87.3\%, Claude~3.7 Sonnet by 78.8\%, and Gemini~2.5~Pro by 87.2\%. Aggregating across all tasks and prompt settings, reasoning-focused models degrade by roughly 20-30\%, while general-purpose models drop by about 45-50\% on average.

\noindent\textbf{Task-Level Vulnerabilities Under Logical Obfuscation.} Obfus Blood Relation shows a steady decline from base to Obfuscation L1 and L2: Claude~3.7 (CoT) drops from 59\% (base) to 24\% (L2), while GPT-5 (few-shot) declines from 73\% to 59\% (–19\%). Obfus Number Series is especially punishing for general-purpose models, GPT-4o falls by 87\%, Claude~3.7 by 79\%, and Gemini~2.5~Pro by 87\%, whereas reasoning-focused models sustain comparatively higher accuracy, even on base questions. Obfus Direction Sense stresses state tracking: GPT-5 (few-shot) drops by 24\%, Qwen~QwQ-32B (few-shot) by 29\%, and Claude~3.7 (zero-shot) by 41\%. These results highlight how logical obfuscation amplifies the cognitive load across tasks, exposing weaknesses in reasoning consistency.

\vspace{-1mm}
\noindent\textbf{Prompting Strategies Provide Partial Relief, Not Robustness.} Advanced prompting methods like CoT reduce GPT-5's average degradation from 27.00\% (zero-shot) to 18.70\%, but the effect remains inconsistent; Gemini~2.5~Pro still suffers drops exceeding 40\% across all prompt settings (cf. Table~\ref{tab:obfuscation_performance_compact}). Few-shot gains are similarly unstable, indicating models' sensitivity to the surface form of examples. While scale and prompting improve base-question accuracy, they fail to ensure invariance to logically equivalent reformulations.

\noindent\textbf{Lower-Parameter Models Exhibit Greater Brittleness.} As shown in Table~\ref{tab:smaller_models_comparison}, smaller models incur sharply higher performance losses. GPT-4o-mini’s accuracy drops by 48.90-51.30\% across prompts, while Gemma-3-27B-IT declines by 39.28-44.86\%. Llama-4-Maverick-17B shows notable prompt sensitivity, its 24.75\% zero-shot drop worsens to over 42\% with few-shot guidance. This fragility is evident in tasks like Obfus Direction Sense, where GPT-4o-mini’s accuracy collapses from 21\% to 5\%. Overall, smaller models remain especially brittle to the state-tracking and symbol-masking challenges introduced by Logifus, even with explicit prompt guidance.
\vspace{-2mm}

\subsection{Mechanistic Analysis of LLMs}
To better understand \textit{why} these LLMs fail under logical obfuscation on LogiQAte tasks, we conduct an intrinsic, model-level analysis focusing on two aspects: first, we perform a \textit{memorization sensitivity analysis} to test whether obfuscated inputs trigger higher dependence on memorized patterns from training data, thereby reducing genuine reasoning, and second, we compute \textit{layer-wise next-token log probabilities} to trace how confidence evolves as the model processes a logically equivalent question. This helps identify where within the network reasoning begins to degrade. All experiments are conducted on LLaMA 3.1 8B~\cite{touvron2023llama} (due to budget constraints) using a single NVIDIA A100 GPU.

\begin{figure}[t]
\centering
\includegraphics[width=\columnwidth]{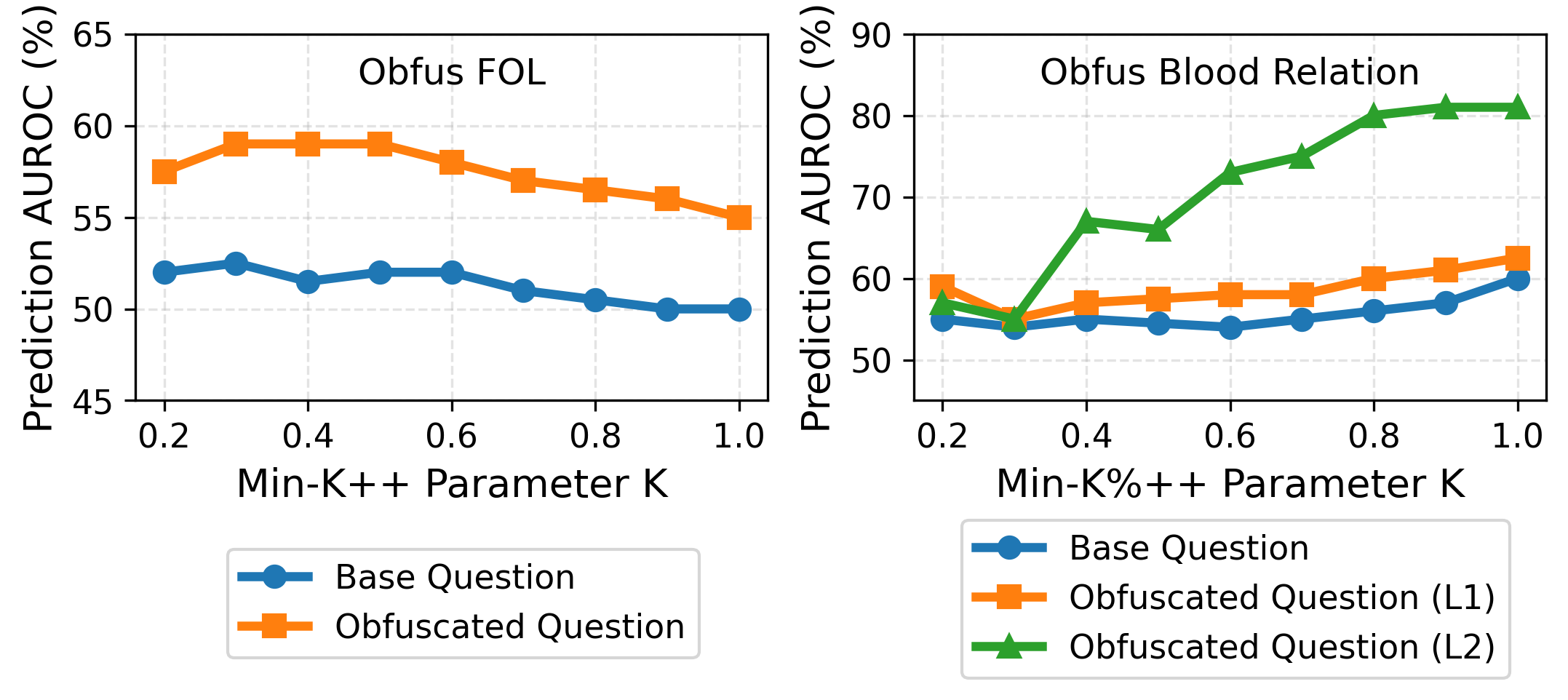}
\caption{
\small Obfuscation increases prediction AUROC (indicating greater memorization) by \textcolor{green!60!black}{12-15\%} in Obfus FOL and \textcolor{green!60!black}{20-25\%} in Obfus Blood Relation across both L1 and L2 levels (L2 being higher). The y-axis shows Prediction AUROC (\%), and the x-axis denotes K, the fraction of least-likely tokens used in computing membership scores. Plotting Prediction AUROC against K illustrates how obfuscation amplifies memorization cues, making models rely more on training data patterns than genuine reasoning.
}

\label{fig:introFig}
\vspace{-6mm}
\end{figure}
\vspace{-2mm}
\paragraph{Memorization Sensitivity Analysis.}
\label{mem}
As shown in Fig.~\ref{mem}, we apply the Min-K\%++~\cite{zhang2025mink} membership inference technique, which detects memorization by checking whether a model assigns unusually high confidence to tokens seen during training. The parameter K controls the fraction of least-likely tokens (by log probability) used when estimating a question’s membership score, lower K focuses on the model’s most uncertain regions, while higher K aggregates over more of the sequence. By plotting prediction AUROC across varying K, we visualize how confidence patterns differ between base and obfuscated questions. A higher AUROC indicates stronger memorization and greater reliance on training data rather than genuine reasoning.

In \text{Obfus FOL}, obfuscated questions consistently yield higher AUROC scores (56-59\%) than their base counterparts (49-53\%) across K, showing increased reliance on memorized logical patterns under obfuscation. In \text{Obfus Blood Relation}, the effect is even stronger: Level 1 obfuscation reaches about 62\% AUROC, while Level 2 rises to 82\% (both peaking at K=1.0). Base questions remain moderate (55-60\%), indicating that relational reasoning tasks involving entity-based relationships make the model more prone to memorization.
 \begin{figure}[htbp]
    \centering
    \includegraphics[width=\linewidth]{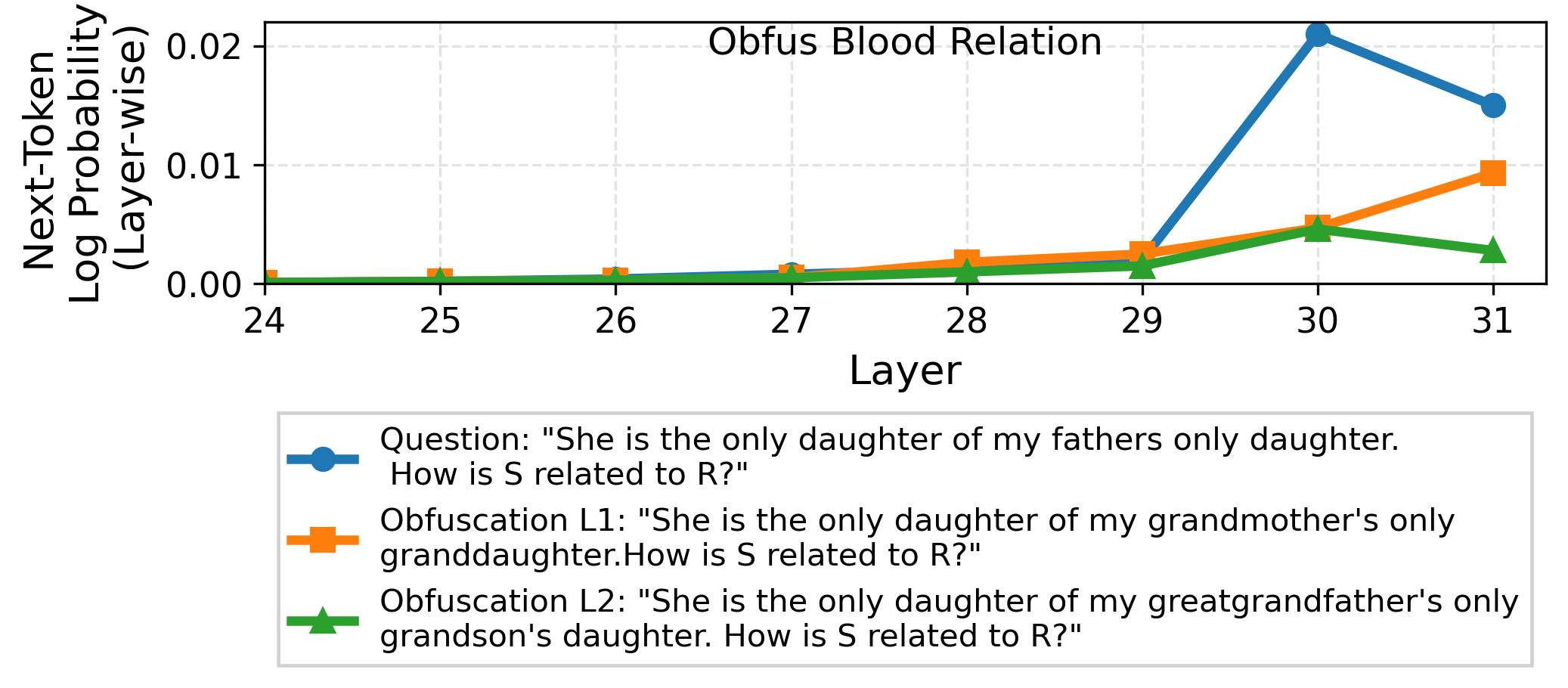}
    \vspace{3mm} 
    \includegraphics[width=\linewidth]{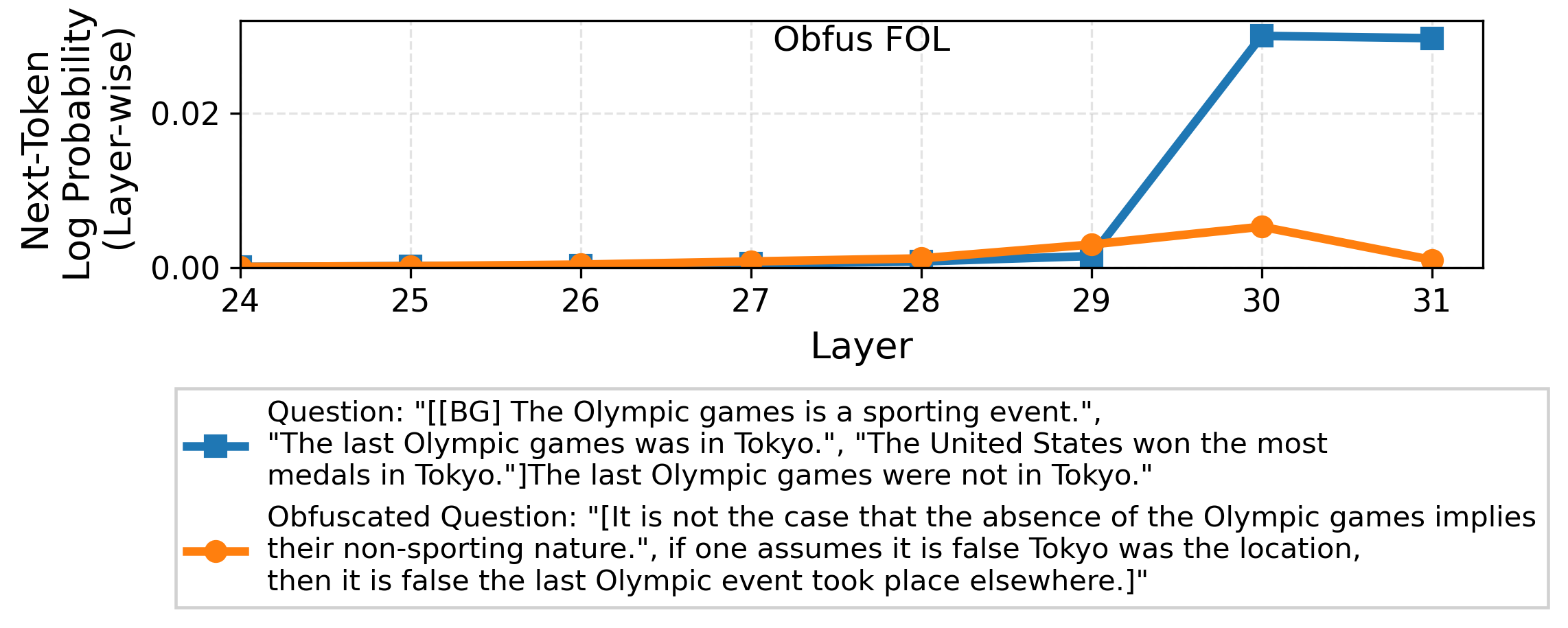}
\vspace{-8mm}
\caption{\small Logical obfuscation reduces late-layer confidence (layers 28-31) across both tasks, showing 50-80\% suppression in next-token log probability compared to base questions. This indicates that obfuscation disrupts the model’s deep-layer reasoning and semantic consolidation.}

    \label{fig:stacked_figures}
    \vspace{-6mm}
\end{figure}
\paragraph{Layer-wise Next Token Confidence Analysis.}
\label{layerwise}
Mid-to-late transformer layers have been shown to capture semantic and relational understanding \cite{Tenney2019, Wei2022}. In Fig.~\ref{fig:stacked_figures}, for each transformer layer \( l \) and target token \( x_t \), we report the layer-wise next-token log probability \( \ell^{(l)}_t = \log p_\theta(x_t \mid x_{<t}, h^{(l)}) \) and the sequence-average \( \bar{\ell}^{(l)} = \frac{1}{T}\sum_{t=1}^T \ell^{(l)}_t \) for both base and obfuscated questions to observe how confidence evolves across layers and whether obfuscation affects deep-layer consolidation. In both of the tasks, the late-layer region (layers 28-31) represents where the model typically consolidates semantic and relational reasoning signals before final prediction. In Obfus Blood Relation, we observe that obfuscated variants (L1 and L2) exhibit a reduction of nearly 50-75\% in overall confidence magnitude than base questions. In Obfus FOL, the obfuscated questions show over 80\% suppression of the confidence rise observed in the base question. These results show that logical obfuscation adversely affects the model’s reasoning ability, as it disrupts the most important deep layers that are primarily responsible for integrating semantic cues and producing the correct answer.

\vspace{-2mm}
\section{Conclusion}
In this work, we introduce \text{Logifus}, a novel structure- and equivalence-preserving logical obfuscation algorithm, and leveraging it, present \text{LogiQAte}, the first benchmark designed to rigorously evaluate the reasoning capabilities of LLMs under logical obfuscation across four complementary tasks. Our findings reveal that logical obfuscation consistently weakens model reasoning, indicating that current LLMs rely heavily on surface-level patterns rather than genuine understanding.
\section{Limitations and Future Work}
While LogiQAte provides a strong foundation for evaluating logical reasoning in large language models, it is currently limited to tasks in English. In future work, we plan to expand the dataset to include multilingual data, with a particular emphasis on low-resource languages. To improve generalizability and diversity, we also aim to extend LogiQAte to additional reasoning categories such as seating arrangements and commonsense inference. Furthermore, developing adaptive logical obfuscation techniques, could strengthen the diagnostic depth and robustness of the framework.

\section*{Ethics Statement}
All paid models used in this study were accessed through valid subscriptions and in full compliance with the terms of service of their respective providers. No attempts were made to bypass usage restrictions or violate the ethical guidelines established by the model developers.

\bibliography{main}

\appendix
\section{Appendix}
\subsection{Performance on More Models}
\label{moremodels}
Table~\ref{tab:more_models_comparison} presents a comprehensive
performance comparison between GPT-4o and Gemini 2.5 Pro across all four
LogiQAte tasks under zero-shot, few-shot, and CoT prompt
setting. Both models demonstrate   an average performance drop of 45.66\% across all
prompting configurations under logical obsfuscation. GPT-4o shows particularly severe degradation in
Obfus Direction Sense tasks (76\% average drop), while Gemini 2.5 Pro, despite
being a reasoning model, achieves near-zero accuracy on Obfus Number
Series tasks (87\% drop). Gemini 2.5's technical report acknowledges
that current reasoning architectures still struggle when benchmark
patterns are disrupted, exposing persistent model blind spots even in
basic logical tasks~\cite{gemini2025v25report}, explaining why even these
reasoning-enhanced models experience substantial
accuracy collapse.
\begin{table*}[t]
\centering
\footnotesize
\setlength{\tabcolsep}{4.2pt}
\renewcommand{\arraystretch}{1.12}
\begin{adjustbox}{max width=\textwidth}
\begin{tabular}{l*{2}{ccc}}
\toprule
\textbf{Task \& Question Variant} &
\multicolumn{3}{c}{\textbf{GPT-4o}} &
\multicolumn{3}{c}{\textbf{Gemini 2.5 Pro}} \\
\cmidrule(lr){2-4}\cmidrule(lr){5-7}
 & \textbf{Zero-shot} & \textbf{Few-shot} & \textbf{CoT}
 & \textbf{Zero-shot} & \textbf{Few-shot} & \textbf{CoT} \\
 & \multicolumn{3}{c}{\textit{(No Reasoning)}} & \multicolumn{3}{c}{\textit{(Reasoning Model)}} \\
\midrule
\multicolumn{7}{l}{\textit{Obfus FOL}} \\
Base        & 0.86 & 0.90 & 0.91 & \cellcolor{red!20}0.78 & \cellcolor{red!20}0.86 & \cellcolor{red!20}0.80 \\
Obfuscation & 0.74 & 0.75 & 0.76 & \cellcolor{red!20}0.68 & \cellcolor{red!20}0.66 & 0.76 \\
\midrule
\multicolumn{7}{l}{\textit{Obfus Blood Relation}} \\
Base        & 0.57 & 0.57 & 0.59 & \cellcolor{red!20}0.51 & \cellcolor{red!20}0.51 & \cellcolor{red!20}0.53 \\
Obfuscation L1    & 0.37 & 0.40 & 0.37 & \cellcolor{red!20}0.35 & \cellcolor{red!20}0.34 & \cellcolor{red!20}0.36 \\
Obfuscation L2    & 0.27 & 0.26 & 0.26 & \cellcolor{red!20}0.19 & \cellcolor{red!20}0.19 & \cellcolor{red!20}0.25 \\
\midrule
\multicolumn{7}{l}{\textit{Obfus Number Series}} \\
Base        & 0.24 & 0.22 & 0.33 & \cellcolor{red!20}0.15 & \cellcolor{red!20}0.18 & \cellcolor{red!20}0.21 \\
Obfuscation & 0.04 & 0.03 & 0.03 & \cellcolor{red!20}0.02 & \cellcolor{red!20}0.02 & \cellcolor{red!20}0.03 \\
\midrule
\multicolumn{7}{l}{\textit{Obfus Direction Sense}} \\
Base        & \cellcolor{red!20}0.29 & \cellcolor{red!20}0.29 & \cellcolor{red!20}0.27 & 0.52 & 0.69 & 0.52 \\
Obfuscation & \cellcolor{red!20}0.14 & \cellcolor{red!20}0.14 & \cellcolor{red!20}0.14 & 0.41 & 0.52 & 0.38 \\
\midrule
\textbf{Average Performance Degradation (\%)} &
\textcolor{red}{-47.60\%} & \textcolor{red}{-47.22\%} & \textcolor{red}{-49.37\%} &
\textcolor{red}{-42.91\%} & \textcolor{red}{-46.33\%} & \textcolor{red}{-40.54\%} \\
\bottomrule
\end{tabular}
\end{adjustbox}
\caption{Performance degradation is observed across four logical
obfuscation tasks in LogiQAte, with average drops ranging from
\textcolor{red}{40.54\%} to \textcolor{red}{49.37\%} compared to base
question performance. GPT-4o operates without reasoning configuration,
while Gemini 2.5 Pro is a reasoning model. Each model is evaluated under
Zero-shot, Few-shot, and Chain-of-Thought (CoT) prompt settings.
\colorbox{red!20}{Red} cells indicate the lowest score for each prompting
strategy within each task row.}
\label{tab:more_models_comparison}
\end{table*}

\subsection{Performance on Smaller Models} 
\label{smallmodels}
Table~\ref{tab:smaller_models_comparison} extends the evaluation to four
additional lower-parameter non-reasoning models: GPT-4o-mini,
Llama-4-Maverick-17B, and Gemma-3-27B-IT and Claude-4.5-Haiku. Performance
degradation averages 42.77\% across all prompt settings, with
GPT-4o-mini experiencing the most severe drops, 50.27\% on an average. All four of these models collapse nearly completely on Obfus Number Series under logical obfuscation, with accuracy falling to near-zero levels representing drops exceeding 78\% from
already-weak baselines.

\begin{table*}[t]
\centering
\footnotesize
\setlength{\tabcolsep}{4.2pt}
\renewcommand{\arraystretch}{1.12}
\begin{adjustbox}{max width=\textwidth}
\begin{tabular}{l*{4}{ccc}}
\toprule
\textbf{Task \& Question Variant} &
\multicolumn{3}{c}{\textbf{GPT-4o-mini}} &
\multicolumn{3}{c}{\textbf{Llama-4-Maverick-17B}} &
\multicolumn{3}{c}{\textbf{Gemma-3-27B-IT}} &
\multicolumn{3}{c}{\textbf{Claude-4.5-Haiku}} \\
\cmidrule(lr){2-4}\cmidrule(lr){5-7}\cmidrule(lr){8-10}\cmidrule(lr){11-13}
 & ZS & FS & CoT & ZS & FS & CoT & ZS & FS & CoT & ZS & FS & CoT \\
 & \multicolumn{12}{c}{\textit{(No Reasoning)}} \\
\midrule
\multicolumn{13}{l}{\textit{Obfus FOL}} \\
Base &
0.73 & 0.57 & 0.56 &
\cellcolor{red!20}0.65 & 0.65 & \cellcolor{red!20}0.45 &
0.82 & 0.80 & 0.78 &
0.86 & 0.88 & 0.53 \\
Obfuscation &
0.73 & 0.57 & 0.56 &
\cellcolor{red!20}0.65 & 0.65 & \cellcolor{red!20}0.45 &
0.71 & 0.73 & 0.62 &
0.67 & 0.72 & 0.67 \\
\midrule
\multicolumn{13}{l}{\textit{Obfus Blood Relation}} \\
Base &
0.50 & 0.46 & 0.48 &
0.25 & 0.41 & \cellcolor{red!20}0.21 &
\cellcolor{red!20}0.11 & 0.49 & 0.46 &
0.50 & 0.54 & 0.39 \\
Obfuscation L1 &
0.29 & \cellcolor{red!20}0.28 & 0.28 &
0.22 & 0.31 & \cellcolor{red!20}0.17 &
\cellcolor{red!20}0.06 & 0.30 & 0.27 &
0.34 & 0.37 & 0.14 \\
Obfuscation L2 &
0.23 & 0.24 & 0.20 &
\cellcolor{red!20}0.13 & \cellcolor{red!20}0.18 & \cellcolor{red!20}0.08 &
0.10 & 0.21 & 0.25 &
0.27 & 0.25 & 0.11 \\
\midrule
\multicolumn{13}{l}{\textit{Obfus Number Series}} \\
Base &
0.24 & 0.22 & 0.33 &
\cellcolor{red!20}0.01 & \cellcolor{red!20}0.02 & \cellcolor{red!20}0.05 &
0.14 & 0.16 & 0.13 &
0.24 & 0.26 & 0.32 \\
Obfuscation &
0.04 & 0.03 & 0.03 &
\cellcolor{red!20}0.01 & \cellcolor{red!20}0.00 & \cellcolor{red!20}0.00 &
0.03 & 0.03 & 0.03 &
0.05 & 0.07 & 0.05 \\
\midrule
\multicolumn{13}{l}{\textit{Obfus Direction Sense}} \\
Base &
\cellcolor{red!20}0.21 & \cellcolor{red!20}0.24 & \cellcolor{red!20}0.21 &
0.14 & 0.35 & 0.29 &
0.31 & 0.31 & 0.28 &
0.26 & 0.35 & 0.23 \\
Obfuscation &
\cellcolor{red!20}0.05 & \cellcolor{red!20}0.07 & \cellcolor{red!20}0.08 &
0.13 & 0.23 & 0.22 &
0.17 & 0.18 & 0.18 &
0.10 & 0.21 & 0.10 \\
\midrule
\textbf{Average Performance Degradation (\%)} &
\textcolor{red}{-51.30} & \textcolor{red}{-48.90} & \textcolor{red}{-50.62} &
\textcolor{red}{-24.75} & \textcolor{red}{-42.64} & \textcolor{red}{-39.11} &
\textcolor{red}{-39.28} & \textcolor{red}{-44.86} & \textcolor{red}{-43.45} &
\textcolor{red}{-48.16} & \textcolor{red}{-43.29} & \textcolor{red}{-34.75} \\
\bottomrule
\end{tabular}
\end{adjustbox}
\caption{Performance degradation is observed across four logical obfuscation tasks in LogiQAte, with average drops ranging from \textcolor{red}{24.75\%} to \textcolor{red}{51.30\%} compared to base question performance. GPT-4o-mini, Llama-4-Maverick-17B, Gemma-3-27B-IT and Claude-4.5-Haiku operate with no reasoning configuration. Each model is evaluated under ZS (Zero-shot), FS (Few-shot), and Chain-of-Thought (CoT) prompt settings. \colorbox{red!20}{Red} cells indicate the lowest score for each prompting strategy within each task row.}
\label{tab:smaller_models_comparison}
\end{table*}

\subsection{Qualitative Analysis}
\label{qualitativeanalysis}
To understand the failure modes of LLMs on obfuscated questions, we performed a qualitative analysis of GPT-5's reasoning trace as it justifies its answer. For each LogiQAte task, we selected two representative examples: one where the model produced the correct answer and one where it failed. We prompted the model in the zero-shot setting to predict its answer along with a brief justification of \textit{why}.
Table~\ref{gpt5_obfuscation_reasoning} and Table~\ref{gpt_5_obfuscation_reasoning_2} and presents these examples with the model's justifications for each obfuscated question.\
We manually analyzed each reasoning justification and applied a two-color highlighting scheme, \textcolor{green!60!black}{green} and \textcolor{red}{red}, to identify correct versus incorrect reasoning tokens or traces.\
In one of the examples in Obfus FOL, the model correctly identifies that \textit{"Legend of Zelda is created by such an entity"} and that Zelda sold over one million copies, but erroneously reverses the direction of implication, treating selling over one million copies as merely necessary for Top 10 inclusion rather than sufficient, incorrectly concluding that \textit{"exceeding one million sales does not suffice to be in the Top 10"}. This represents a failure in parsing complex logical negations and understanding implication directionality.\
In Obfus Blood Relation, GPT-5 correctly traces \textit{"my great-grandfather's only grandson"} to V's father and simplifies to the \textit{"only son of my father"}, but then makes an unjustified assumption that V is the \text{only} son, concluding that the boy being pointed at is the \textit{"son of V"}, an error arising from over-interpreting \textit{"only son"} at the wrong generation level.\
The Obfus Number Series incorrect example reveals that GPT-5 correctly decodes the MD5 hashes, identifies the pattern (sum of consecutive squares $1^2, 2^2, 3^2, \ldots$), and determines \textit{"the hidden term between 15 and 56 is 31"}, yet outputs 141 by continuing the sequence \text{beyond} the given terms ($92 + 7^2 = 141$), answering incorrectly.\
In Obfus Direction Sense, the error stems from assuming \textit{"they start facing North"} and \textit{"both walk 1 km North together"} despite the problem stating \textit{"in opposite directions"}. While subsequent turning patterns are correctly identified, this faulty initial condition propagates through all position calculations, yielding 4 km instead of 3 km.

\subsection{Human Annotation Procedure and Instructions}

To ensure the reliability and validity of \text{LogiQAte}, all base and obfuscated questions underwent rigorous human verification. Two independent human annotators, both third-year undergraduate Computer Science students with strong backgrounds in logic, discrete mathematics, and reasoning, were recruited for the validation process. Their familiarity with formal logic systems and demonstrated proficiency in analytical reasoning made them well-suited for this task.

  \textbf{Annotation Protocol.} The two annotators were provided with detailed task-specific guidelines for each reasoning task:

  \begin{itemize}
      \item \textbf{Obfus FOL:} Verify that obfuscated premises remain logically equivalent to base premises and preserve the conclusion's truth value. For example, consider the base premise: \textit{``Humans
  are mammals''} ($\forall x$ $(Human(x) \rightarrow Mammal(x))$). The obfuscated version reads: \textit{``It is false that if there exists an entity that is not an animal, then there exists an entity that is
  a human who is not a mammal''}. Annotators verified that both versions express the same logical relationship and that any conclusion derived from the base premises holds for the obfuscated premises. They
  flagged instances where logical transformations altered the truth value of the conclusion.

      \item \textbf{Obfus Blood Relation:} Confirm that multi-hop relational chains preserve the final relationship. For instance, the base statement \textit{``D is the wife of C, C is the father of F''}
  establishes that D is F's mother. The obfuscated version states: \textit{``D is the wife of C, C is the grandfather's only son of F''}. Annotators traced the obfuscated chain step-by-step: if C is the
  grandfather's only son of F, then C must be F's father, thus D remains F's mother. They verified that despite the indirect phrasing, the underlying family graph and final answer remain unchanged.

      \item \textbf{Obfus Number Series:} Validate that symbolic substitutions preserve the underlying mathematical pattern. For example, the base sequence $[2, 4, 6, 8]$ follows an arithmetic progression
  (+2). Under Type 1 obfuscation, this becomes $\{\text{Venus}, \text{Mars}, \text{Jupiter}, \text{Saturn}\}$ using the planet mapping (Venus=2, Mars=4, Jupiter=6, Saturn=8). Annotators confirmed that (1) the
  mapping is applied correctly, (2) the arithmetic pattern remains recoverable, and (3) the next term can be uniquely determined by decoding the planet names back to numbers.

      \item \textbf{Obfus Direction Sense:} Verify that added detours are self-canceling and preserve net displacement. Consider the base path: \textit{``A person walks 5 km north and then 3 km east''} (final
  position: 5.83 km North-East). The obfuscated version states: \textit{``A person walks 6 km north, then 4 km east, 6 km south, 3 km east, 4 km west, and finally 5 km north''}. Annotators computed each
  segment: the 6 km north and 6 km south cancel out, as do the 4 km east and 4 km west, leaving the net movement of 5 km north and 3 km east, identical to the base path. They verified that all inserted
  movements form canceling pairs and the final displacement remains 5.83 km North-East.
  \end{itemize}

Both annotators worked independently, and disagreements were resolved through discussion until consensus was reached. For \text{Obfus FOL}, the automated theorem prover,  \text{Prover9}~\cite{prover9-mace4} was used as an additional verification step prior to human review.

\textbf{Inter-Annotator Agreement.} We measured inter-annotator reliability using Cohen’s kappa ($\kappa = \frac{p_o - p_e}{1 - p_e}$), which accounts for chance agreement between the two human annotators. The results demonstrated strong consistency across all tasks: \text{Obfus FOL} achieved $\kappa = 0.95$ for base question validation and $\kappa = 0.89$ for obfuscated questions; \text{Obfus Blood Relation} obtained $\kappa = 0.90$; and \text{Obfus Direction Sense} reached $\kappa = 0.88$. Overall, the benchmark achieved near-perfect agreement across all tasks ($\kappa > 0.88$), confirming that logical equivalence and answer preservation were reliably maintained throughout the obfuscation process. This high agreement underscores both the clarity of our annotation guidelines and the logical soundness of the \text{Logifus} framework.

\begin{table*}[t]
\centering
\caption{Qualitative Analysis of GPT-5 on LogiQAte: Reasoning traces showing correct answers and reasoning steps in 
\textcolor{green!60!black}{green} and incorrect ones (answers and reasoning steps) in \textcolor{red}{red} across four LogiQAte tasks. We observe that, even when GPT-5 arrives at an incorrect answer, portions of its reasoning are often correct; the model typically makes localized errors at specific inference steps rather than failing completely.}
\label{gpt5_obfuscation_reasoning}
\begin{adjustbox}{max width=\textwidth}
\begin{tabular}{|>{\raggedright\arraybackslash}p{3cm}|>{\raggedright\arraybackslash}p{4.5cm}|>{\raggedright\arraybackslash}p{4.5cm}|c|>{\raggedright\arraybackslash}p{7.5cm}|}
\hline
\textbf{Task} & \textbf{Base Question} & \textbf{Obfus Question} & \textbf{Obfus Answer} & \textbf{Reasoning} \\
\hline

\multirow{2}{=}{\parbox{3cm}{\centering\vspace{7cm}\textbf{Obfus FOL}: Given premises and conclusion, determine if conclusion logically follows it. Output TRUE or FALSE with brief justification.}}
&

Diamond Mine is a professional wrestling stable, formed in WWE. Roderick Strong leads Diamond Mine. Diamond Mine includes the Creed Brothers, and Ivy Nile. Imperium has a feud with Diamond Mine.

\textbf{Ground Truth:} FALSE
&

It is not the case that if Diamond Mine is not a professional wrestling stable or it was not formed in WWE, then a contradiction arises. If it is not the case that Roderick Strong does not lead Diamond Mine, then Roderick Strong leads, or an impossibility occurs. If Diamond Mine includes the Creed Brothers, it is not not the case that it includes Ivy Nile as well. It is not that Imperium does not avoid having a feud with Diamond Mine, unless there exists a case where Imperial peace is violated.
&
\textcolor{green!60!black}{\textbf{FALSE}}
&
The conclusion claims that ``Imperium doesn't have a feud with a professional wrestling stable that includes Ivy Nile'', which in formal terms requires that for every professional wrestling stable including Ivy Nile, Imperium has no feud with it. \hl{However, the premises never establish such a universal restriction. The third premise only states that if Diamond Mine includes the Creed Brothers, then it also includes Ivy Nile, which is a conditional relationship about inclusion and does not mention feuds at all.} The fourth premise, \hl{``It is not the case that Imperium does not avoid having a feud with Diamond Mine...''} is logically weak and \hl{still allows the existence of feuds}, especially since its formal version permits $\exists y \text{HasFeud}(\text{imperium}, y)$. Likewise, the \hl{first and second premises focus on internal logical consistency, leadership, or impossibilities involving Diamond Mine and Roderick Strong}, but \hl{they do not rule out Imperium having a feud with a stable that includes Ivy Nile.} \hl{Since the conclusion requires a global negation of feuds across all such stables, and the premises neither assert nor imply this universal condition, the conclusion does not logically follow.} Therefore, the correct label is False. \\
\cline{2-5}

&
A Japanese game company created the game the Legend of Zelda. All games in the Top 10 list are made by Japanese game companies. If a game sells more than one million copies, then it will be selected into the Top 10 list. The Legend of Zelda sold more than one million copies.

\textbf{Ground Truth:} TRUE
&
It is not the case that the absence of a Japanese entity responsible for video games results in the Legend of Zelda being unlinked with creation by such an entity. The state where a Japanese creator is absent implies that it's not possible for a game created and present in the Top 10 list to exist. For every game, the condition of surpassing sales of a million copies is a necessity for it not to be excluded from the Top 10 list. It is not the case that the Legend of Zelda didn't exceed selling more than one million copies.
&
\textcolor{red}{\textbf{FALSE}}
&
From the denial of `if no Japanese entity exists then the Legend of Zelda is not created by one,' it follows that \hlred{the absence of a Japanese creator holds} and, nonetheless, \hl{the Legend of Zelda is created by such an entity}. Given that absence, \hlred{the second premise says that no game that is created and in the Top 10 can exist}; in particular, the Legend of Zelda cannot be in the Top 10. The \hlred{sales premises add only a necessary condition: any Top 10 game must have sold over one million copies}, and indeed \hl{Zelda did}, but \hlred{exceeding one million sales does not suffice to be in the Top 10}. \hlred{Hence the conclusion that the Legend of Zelda is in the Top 10 does not follow and is in fact contradicted.} \\

\hline

\multirow{2}{=}{\parbox{3cm}{\centering\vspace{4cm}\textbf{Obfus Blood Relation}: Given a relationship description, determine the final relationship with brief
justification.}}
&
Pointing towards a boy V said He is the son of only son of my grandfather, How is that boy related to V?

\textbf{Ground Truth:} BROTHER
&
Pointing towards a boy V said He is the son of only son of my father's father, How is that boy related to V?
&
\textcolor{green!60!black}{\textbf{BROTHER}}
&
\hl{My father's father'' is V's grandfather.} \hl{The only son'' of V's grandfather must be V's father} (since the grandfather has exactly one son). So the phrase becomes: \hl{He is the son of my father.''} \hl{A male child of V's father is V's brother.} \hl{It is not V himself because V is pointing to another boy as he,''} and standard reasoning assumes the speaker is not indicating themselves. \hl{Therefore, the boy pointed out is V's brother.} \\

\cline{2-5}

&
Pointing towards a boy V said He is the son of only son of my grandfather, How is that boy related to V?

\textbf{Ground Truth:} BROTHER
&
Pointing towards a boy V said He is the son of only son of my greatgrandfather's only grandson, How is that boy related to V?
&
\textcolor{red}{\textbf{Son}}
&
\hl{My great-grandfather's only grandson'' is two generations below and thus in the parent's generation;} with only one male grandchild, that person is V's father. \hl{The phrase becomes: the only son of my father,''} \hlred{which identifies the single male child of V's father. Since there is exactly one such son, that son is V himself} (V is the only son of his father). \hlred{Therefore, the boy pointed at is ``the son of [V],'' i.e., V's son.} \hlred{So, the boy is V's son.} \\

\hline

\end{tabular}
\end{adjustbox}
\end{table*}

\begin{table*}[t]
\centering
\caption{(Continued) Qualitative Analysis of GPT-5 on LogiQAte: Reasoning traces showing correct answers and reasoning steps in \textcolor{green!60!black}{green} and incorrect ones (answers and reasoning steps) in \textcolor{red}{red} across four LogiQAte tasks. We observe that, even when GPT-5 arrives at an incorrect answer, portions of its reasoning are often correct; the model typically makes localized errors at specific inference steps rather than failing completely.}
\label{gpt_5_obfuscation_reasoning_2}
\begin{adjustbox}{max width=\textwidth}
\begin{tabular}{|>{\raggedright\arraybackslash}p{3cm}|>{\raggedright\arraybackslash}p{4.5cm}|>{\raggedright\arraybackslash}p{4.5cm}|c|>{\raggedright\arraybackslash}p{7.5cm}|}
\hline
\textbf{Task} & \textbf{Base Question} & \textbf{Obfus Question} & \textbf{Obfus Answer} & \textbf{Reasoning} \\
\hline

\multirow{2}{=}{\parbox{3cm}{\centering\vspace{1cm}\textbf{Obfus Number Series}: Given a number series with planets representing digits or hashes representing single digits, find the missing term. \\
\text{Planet Mapping:} Sun:0, Mercury:1, Venus:2, ..., Neptune:8\\
\text{Hash Mapping:} cfcd...
:0, c4ca...
:1, c81e...
:2, ..., 45c4...
:9}}
&
2048, 516, 1032, 262, ?, 135, 270
\textbf{Ground Truth:} 524
&
2048, Jupiter Mercury Saturn, Mercury Sun Earth Venus, Venus Saturn Venus, ?, Mercury Earth Jupiter, Venus Uranus Sun
&
\textcolor{green!60!black}{\textbf{524}}
&
Using the code \hl{Sun=0, Mercury=1, Venus=2, Earth=3, Mars=4, Jupiter=5, Saturn=6, Uranus=7, Neptune=8, Pluto=9}, the planets translate to digits: \hl{Jupiter Mercury Saturn $\rightarrow$ 516, Mercury Sun Earth Venus $\rightarrow$ 1032, Venus Saturn Venus $\rightarrow$ 262, Mercury Earth Jupiter $\rightarrow$ 135, Venus Uranus Sun $\rightarrow$ 270.} So the numeric sequence is \hl{2048, 516, 1032, 262, ?, 135, 270.} \hl{The terms follow an alternating rule: (1) divide by 4 and add 4; (2) multiply by 2.} Check: 2048/4+4=516; 516$\times$2=1032; 1032/4+4=262; then 262$\times$2=524; 524/4+4=135; 135$\times$2=270. \hl{Hence the missing term is 524.} \\

\cline{2-5}

&
1, 2, 6, 15, ..., 56, 92

\textbf{Ground Truth:} 31
&
{\scriptsize 1, c81e728d9d4c2f636f067f89cc14862c, 1679091c5a880faf6fb5e6087eb1b2dc, 15, ..., 56, 92}
&
\textcolor{red}{\textbf{141}}
&
Decode the hashes as MD5 of single digits: \hl{c81e728d9d4c2f636f067f89cc14862c = 2 and 1679091c5a880faf6fb5e6087eb1b2dc = 6.} So the visible sequence is \hl{1, 2, 6, 15, ..., 56, 92.} \hl{The differences form 1, 4, 9, 16, 25, 36, i.e., consecutive squares $1^2$ through $6^2$.} \hl{Equivalently, $a_n = a_{n-1} + n^2$ with $a_1 = 1$.} \hl{Thus the hidden term between 15 and 56 is 31 (15 + 16).} \hlred{Continuing, the next increment is $7^2 = 49$, giving 92 + 49 = 141.} MD5('2') and MD5('6') are standard, widely known values, confirming the intended decoding. Summing consecutive squares accumulates as 1, 1+4, 1+4+9, ... producing 1, 2, 6, 15, 31, 56, 92. \hlred{Each step adds the next square in order.} \\

\hline

\multirow{2}{=}{\parbox{3cm}{\centering\vspace{1.5cm}\textbf{Obfus Direction Sense}: Given a navigation path with conditional turns, determine the final position with a brief justification.}}
&
From point A, a traveler moves 5 km North. From there, he proceeds 3 km East. Relative to point A, what is the final direction and distance of the traveler?

\textbf{Ground Truth:} 5.83 km away, North-East
&
A man walks 6 km North. If he sees a shop, he must turn right; if not, he must turn left. He sees a shop, so he turns right and goes 4 km. Then, as per his rule, he again turns right and walks 6 km. Finally, since he sees a park, he turns left and walks 3 km. He finds a wall there and comes back for 4 km. At that spot, he looks again. This time he does not see a shop, so he turns right and walks 5 km. Where is he now?
&
\textcolor{green!60!black}{\textbf{5.83 km away, North-East}}
&
Step-by-step (taking right/left relative to current heading): \hl{Start at (0,0). Walk 6 km North to (0,6).} He sees a shop, so \hl{turn right (facing North $\rightarrow$ right is East) and go 4 km to (4,6).} As per the rule, he again \hl{turns right (East $\rightarrow$ right is South) and walks 6 km to (4,0).} Seeing a park, \hl{he turns left (South $\rightarrow$ left is East) and goes 3 km to (7,0).} Finding a wall, he \hl{comes back 4 km along the same path (West) to (3,0).} There he does not see a shop, but as stated \hl{he turns right (West $\rightarrow$ right is North) and walks 5 km to (3,5).} \hl{Net displacement: 3 km East and 5 km North; distance = $\sqrt{3^2+5^2} \approx$ 5.83 km, direction North-East.} \\

\hline

&
Shahid and Rohit start from the same point in opposite directions. After each 1 km, Shahid always turns left, Rohit always turns right. When do they first meet?

\textbf{Ground Truth:} 3 km
&
Shahid and Rohit meet at sunrise at the old banyan tree. After each 1 km, Shahid always turns left, Rohit always turns right. Both walk 1 km. They glance at the fields. They turn, and walk 1 km. A child waves from a bicycle. Turn again and walk 1 km. They pass a vendor selling sugarcane. Turn and walk 1 km. A dog barks; they both stop to shoo it away (no walking during the stop). Turn and walk 1 km. They reach a small pond; Shahid tosses a pebble in, Rohit counts ripples. Turn and walk 1 km. They pass an old woman knitting by the roadside. Turn and walk 1 km. A milkman slows them to ask directions (no extra distance). Turn and walk 1 km. They cross a narrow bridge; Rohit hears church bells. After how many 1-km moves (i.e., after how many kilometers walked by each) do Shahid and Rohit first meet?
&
\textcolor{red}{\textbf{4 km}}
&
\hlred{Assume they start facing North. Move 1: both walk 1 km North together}; they haven't separated yet. Thereafter \hl{Shahid always turns left and Rohit always turns right after each kilometer.} \hl{Shahid's sequence of headings cycles N $\rightarrow$ W $\rightarrow$ S $\rightarrow$ E $\rightarrow$ N, while Rohit's is N $\rightarrow$ E $\rightarrow$ S $\rightarrow$ W $\rightarrow$ N.} Positions after each move from the origin (0,0): \hlred{after 1 km both at (0,1). After 2 km: Shahid at (-1,1), Rohit at (1,1). After 3 km: Shahid (-1,0), Rohit (1,0). After 4 km: both return to (0,0), the starting point.} \hlred{This is their first meeting after they have turned in opposite directions.} \hl{The intermediate incidents add no distance}, so \hlred{the first re-meeting is after 4 km.} \\

\hline

\end{tabular}
\end{adjustbox}
\end{table*}

\clearpage
\newpage
\onecolumn
\subsection{Logifus Algorithms}
\label{logifusalgos}
We list the algorithms (prompts and hardcoded programs) for all four reasoning tasks in \text{Logifus}, thereby yielding \text{LogiQAte} after logical obfuscation. Each algorithm is processed through \text{GPT-4o} with a generation temperature of 0.0 to ensure deterministic outputs. A subsequent review process involves manual verification by two independent human reviewers. For \text{Obfus FOL}, each generated instance is first validated using the \text{Prover9}~\cite{prover9-mace4} theorem prover to confirm logical equivalence, followed by verification by the same two human reviewers to ensure correctness.

\begin{llmprompt}[Prompt used for Simplifying FOLIO Questions in Obfus FOL]

\vspace{1em}

\#\# Task:
Transform a FOLIO dataset row into a simplified logical reasoning problem with a "True" or "False" label.

\#\# Input Structure:
Premises NL: {premises nl}  

Premises FOL: {premises fol}  

Conclusion NL: {conclusion nl}  

Conclusion FOL: {conclusion fol}  

Label: {label}

\vspace{1em}

\#\# Selection Rules:

1. Only process rows where the label is "True" or "False" (exclude any "Uncertain" cases)  

2. Select any two premises from Premises NL that form a logical relationship with each other  

3. Choose any one statement from the remaining Premises NL or use the original Conclusion NL as your new conclusion  

4. Extract the corresponding formulas from Premises FOL and Conclusion FOL for your selections

\vspace{1em}

\#\# Simplification Guidelines:

1. Rewrite each chosen premise and conclusion to be brief, clear, shorter than the original, and easy to understand  

2. Preserve all domain-specific terms and naming conventions  

3. Focus on making the logical relationship easy to understand  

4. Model the style after simple examples like: "Tanya is older than Eric. Cliff is older than Tanya."

\vspace{1em}

\#\# Output Format:

1. Two simplified natural language premises labeled as "Premise 1:" and "Premise 2:"  

2. One simplified natural language conclusion labeled as "Conclusion:"  

3. The question: "If the first two statements are true, is the third statement true or false?"  

4. A composite FOL representation: ([Premise1 FOL]) $\land$ ([Premise2 FOL]) $\rightarrow$ ([Conclusion FOL])  

5. The final label: "True" or "False"

\vspace{1em}

\#\# Label Determination:

- The output label should reflect whether the chosen conclusion logically follows from the two premises  

- Aim to preserve the original logical relationship implied by the dataset when possible

\vspace{1em}

Produce only the requested output with no additional explanations.
\vspace{1em}
\end{llmprompt}

\begin{llmprompt}[Logical Obfuscation Prompt used in Obfus FOL]

  \vspace{0.5em}
  You are an expert in logical obfuscation. Your task is to take the input dataset row and create a complex, logically obfuscated version of the premises while preserving the semantic meaning and ground truth
  relationship with the conclusion.

  \vspace{0.5em}

  Input: Premises (Natural Language): {premises}; Premises (FOL): {premises\_fol}; Conclusion (Natural Language): {conclusion}; Conclusion (FOL): {conclusion\_fol}; Label: {label}\\

  \vspace{0.5em}

  Core Requirements:\\
  1. Obfuscate ONLY the premises - do NOT modify the conclusion.\\
  2. Preserve the "True" or "False" label relationship between the obfuscated premises and the original conclusion.\\
  3. Maintain ALL domain-specific vocabulary (words like "employee," "customer," "meeting," etc.).\\
  4. Keep ALL proper names and entity references (like "James," "Bonnie," etc.) exactly as they appear.\\
  5. Transform ONLY the logical structure and connectors using highly nested and compound logical transformations.\\

  \vspace{0.5em}
\end{llmprompt}

\begin{llmprompt}[Logical Obfuscation Prompt used in Obfus FOL - Continued]
  \vspace{1em}

  Logical Transformation Methods (apply at least FOUR per premise, chaining and nesting them together):

 \vspace{1em}

  - Contraposition ("if P then Q" $\rightarrow$ "if not Q then not P"); Double negation ("P" $\rightarrow$ "not not P").\\
  - Logical equivalences ("P or Q" $\rightarrow$ "not(not P and not Q)"); Conditional to disjunction ("if P then Q" $\rightarrow$ "not P or Q").\\
  - De Morgan's laws ("not (P and Q)" $\rightarrow$ "not P or not Q"); Quantifier replacement ("$\forall$x P(x)" $\rightarrow$ "$\neg$$\exists$x $\neg$P(x)").\\
  - Biconditional expansion ("P $\leftrightarrow$ Q" $\rightarrow$ "(P $\rightarrow$ Q) $\land$ (Q $\rightarrow$ P)"); Distribution ("P $\land$ (Q $\lor$ R)" $\rightarrow$ "(P $\land$ Q) $\lor$ (P $\land$ R)").\\
  - Absorption ("P $\lor$ (P $\land$ Q)" $\rightarrow$ "P"); Conditional as conjunction ("P $\rightarrow$ Q" $\rightarrow$ "$\neg$(P $\land$ $\neg$Q)").\\
  - Negation Normal Form ("$\neg$(P $\rightarrow$ Q)" $\rightarrow$ "P $\land$ $\neg$Q"); Quantifier commutation ("$\forall$x $\forall$y P(x,y)" $\rightarrow$ "$\forall$y $\forall$x P(x,y)").\\
  - Redundant tautology ("P" $\rightarrow$ "P $\land$ (Q $\lor$ $\neg$Q)"); Nested compound transformations; Introduce intermediate statements.\\

  \vspace{1em}

  FOL Transformation Guidelines: Maintain all predicate names (e.g., Meeting, LunchInCompany) and constants (e.g., bonnie) exactly as they appear. Do NOT rename or replace any predicates or constants with
  symbols. Only modify the logical connectives ($\rightarrow$, $\land$, $\lor$, $\neg$, $\forall$, $\exists$) and their arrangement. Ensure strict logical equivalence between original and obfuscated FOL. Use multiple nested logical equivalences per
  premise to strongly obscure the original logical structure.\\

  \vspace{1em}

  Output Format:\\

 \vspace{1em}
 
  Obfuscated Premises (Natural Language): ["〈Obfuscated NL premise 1〉", "〈Obfuscated NL premise 2〉", …]\\
  Obfuscated Premises (FOL): ["〈Obfuscated FOL premise 1〉", "〈Obfuscated FOL premise 2〉", …]\\
  Conclusion (Natural Language): {conclusion}; Conclusion (FOL): {conclusion\_fol}; Label: {label}\\

  \vspace{1em}

  Remember: The obfuscation should deeply mask the logical reasoning path, making it significantly less obvious while strictly maintaining logical equivalence and the original truth-value relationship with the
   conclusion. Produce ONLY the exact output format specified above with no introductory text, explanations, reasoning, or comments before or after the requested output. 
   
\vspace{1em}

\end{llmprompt}

\begin{llmprompt}[Logical Obfuscation Algorithm used in Obfus Number Series (Hardcoded)]
  \vspace{1em}

  \textbf{Input:} Base numerical sequence $S = \{n_1, n_2, \ldots, n_k\}$ with underlying pattern.

  \vspace{1em}

  \textbf{Planet Mapping:} $\mathcal{P}$: Sun $\to$ 0, Mercury $\to$ 1, Venus $\to$ 2, Earth $\to$ 3, Mars $\to$ 4, Jupiter $\to$ 5, Saturn $\to$ 6, Uranus $\to$ 7, Neptune $\to$ 8, Pluto $\to$ 9.

  \vspace{1em}

  \textbf{Output:} Three obfuscated sequences $S_1, S_2, S_3$ with preserved pattern.

  \vspace{1em}

  \textbf{Type 1: Planet-Name Substitution}

  \vspace{0.5em}

  - Initialize $S_1 \gets \emptyset$.\\
  - For each $n_i \in S$, add $\mathcal{P}[n_i]$ to $S_1$.\\
  - \textbf{Return:} $S_1 = \{\mathcal{P}[n_1], \mathcal{P}[n_2], \ldots, \mathcal{P}[n_k]\}$.\\
  - \textit{Example:} $[2, 4, 6, 8] \to \{\text{Venus}, \text{Mars}, \text{Jupiter}, \text{Saturn}\}$.\\

  \vspace{1em}

  \textbf{Type 2: ASCII Sum Substitution}

  \vspace{0.5em}

  - Initialize $S_2 \gets \emptyset$.\\
  - For each $n_i \in S$: $\text{planet} \gets \mathcal{P}[n_i]$; $\text{ascii\_sum} \gets \sum_{c \in \text{planet}} \text{ASCII}(c)$; $S_2 \gets S_2 \cup \{\text{ascii\_sum}\}$.\\
  - \textbf{Return:} $S_2 = \{f_{\text{ASCII}}(\mathcal{P}[n_1]), \ldots, f_{\text{ASCII}}(\mathcal{P}[n_k])\}$.\\
  - \textit{Example:} $\text{Mars} \to 77 + 97 + 114 + 115 = 403$.\\

  \vspace{1em}

  \textbf{Type 3: MD5 Hash Substitution}

  \vspace{0.5em}

  - Initialize $S_3 \gets \emptyset$.\\
  - For each $n_i \in S$: $\text{hash} \gets \text{MD5}(\text{str}(n_i))$; $S_3 \gets S_3 \cup \{\text{hash}\}$.\\
  - \textbf{Return:} $S_3 = \{\text{MD5}(\text{str}(n_1)), \ldots, \text{MD5}(\text{str}(n_k))\}$.\\
  - \textit{Example:} $4 \to \texttt{a87ff679a2f3e71d9181a67b7542122c}$.\\

  \vspace{1em}

  \textbf{Dataset Construction:}

  \vspace{0.5em}

  - Collect base sequences from standard aptitude sources.\\
  - For each base sequence $S$ with answer $a$: generate $S_1, S_2, S_3$; verify recoverability; add $(S, a), (S_1, a), (S_2, a), (S_3, a)$ to $\mathcal{D}$.\\
  - \textbf{Return:} Dataset $\mathcal{D}$ containing base and obfuscated sequence pairs.\\

  \vspace{1em}

\end{llmprompt}

\begin{llmprompt}[Logical Obfuscation Prompt used in Obfus Blood Relation (Level 1 and Level 2)]
  \vspace{1em}
  You are a blood relation puzzle assistant.

  \vspace{1em}

  Below are the substitution rules you must follow when modifying the puzzle:

  \vspace{0.5em}

  \textbf{Level 1 Substitutions:}\\
  Father $\rightarrow$ mother's husband, grandfather's only son\\
  Mother $\rightarrow$ Father's wife, grandfather's only daughter-in-law\\
  Brother $\rightarrow$ sister-in-law's husband\\
  Sister $\rightarrow$ Brother-in-law's wife\\
  niece $\rightarrow$ Brother's daughter, sister-in-law's daughter, Father's granddaughter\\
  nephew $\rightarrow$ Brother's son, sister-in-law's son, Father's grandson\\
  Aunt $\rightarrow$ Father's sister, Grandfather's daughter, mother's sister-in-law\\
  Uncle $\rightarrow$ Father's brother, Grandfather's son, mother's brother-in-law\\
  Maternal Uncle $\rightarrow$ mother's brother, Father's brother-in-law, maternal grandmother's son\\
  Paternal Uncle $\rightarrow$ Father's brother, cousin's father\\
  Maternal Aunt $\rightarrow$ mother's sister\\
  Sister-in-law $\rightarrow$ Brother's wife, niece's mother\\
  Brother-in-law $\rightarrow$ Sister's husband\\
  Grand Father $\rightarrow$ father's father, mother's father-in-law\\
  Grand Mother $\rightarrow$ Father's mother, mother's mother-in-law\\
  Daughter-in-law $\rightarrow$ son's wife\\
  Son-in-law $\rightarrow$ Daughter's husband\\
  cousin sister $\rightarrow$ uncle's daughter\\
  cousin brother $\rightarrow$ aunt's son\\
  son $\rightarrow$ Father's grandson\\
  daughter $\rightarrow$ mother's granddaughter\\
  grandson $\rightarrow$ son's son\\
  granddaughter $\rightarrow$ son's daughter\\

  \vspace{1em}

\end{llmprompt}

\begin{llmprompt}[Logical Obfuscation Prompt used in Obfus Blood Relation (Level 1 and Level 2) - Continued]

  \textbf{Level 2 Substitutions:}\\
  Father $\rightarrow$ greatgrandfather's only grandson\\
  Mother $\rightarrow$ greatgrandfather's only grandson's wife\\
  Brother $\rightarrow$ greatgrandfather's only grandson's daughterinlaw's husband\\
  Sister $\rightarrow$ greatgrandfather's only grandson's daughter\\
  niece $\rightarrow$ Father's granddaughter\\
  nephew $\rightarrow$ Father's grandson\\
  Aunt $\rightarrow$ Greatgrandfather's granddaughter, Grandfather's daughter\\
  Uncle $\rightarrow$ greatgrandfather's only grand daughter inlaw's brother inlaw\\
  Maternal Uncle $\rightarrow$ maternal grandmother's son\\
  Maternal Aunt $\rightarrow$ greatgrandfather's only grand daughter inlaw's sister\\
  Sister inlaw $\rightarrow$ greatgrandfather's only grandson's daughterinlaw\\
  Brother inlaw $\rightarrow$ greatgrandfather's only grandson's daughter's husband\\
  Grand Father $\rightarrow$ greatgrandfather's only son\\
  Grand Mother $\rightarrow$ greatgrandfather's only daughter inlaw\\
  Daughter inlaw $\rightarrow$ father inlaw's son's wife\\
  Son inlaw $\rightarrow$ Daughter's husband\\
  cousin sister $\rightarrow$ uncle's daughter\\
  cousin brother $\rightarrow$ aunt's son\\
  son $\rightarrow$ Father's grandson\\
  daughter $\rightarrow$ mother's granddaughter\\
  grandson $\rightarrow$ son's son\\
  granddaughter $\rightarrow$ son's daughter\\

  \vspace{1em}

  Your task:\\
  1. Carefully read the puzzle below.\\
  2. Identify exactly one blood relation word that appears in the puzzle and is listed above.\\
  3. Choose one of the allowed substitutions for that word.\\
  4. Replace that relation word in the puzzle with the selected substitution — only once.\\
  5. Print the modified puzzle, do not add header and footer or any other explanation.\\

  \vspace{1em}

  Puzzle:\\
""" \{puzzle\} """\\

\end{llmprompt}

\begin{llmprompt}[Logical Obfuscation Prompt used in Obfus Direction Sense]
  \vspace{1em}

  Direction sense : You are a direction-sense puzzle maker.

  \vspace{1em}

  Below are the transformation rules you must follow when modifying the puzzle:

  \vspace{1em}

  1. Direction Substitution Rules:\\
     - North $\rightarrow$ facing 12 o'clock\\
     - South $\rightarrow$ facing 6 o'clock\\
     - East  $\rightarrow$ facing 3 o'clock\\
     - West  $\rightarrow$ facing 9 o'clock\\

  \vspace{1em}

  2. Rotation Substitution Rules:\\
     - "turn right"  $\rightarrow$ "rotate 1 quarter-turn clockwise"\\
     - "turn left"   $\rightarrow$ "rotate 1 quarter-turn counterclockwise"\\
     - "turn back"   $\rightarrow$ "rotate 2 quarter-turns"\\
     - "faces opposite direction" $\rightarrow$ "rotates 2 quarter-turns"\\

  \vspace{1em}

  3. Apply Zigzag Cancellation Rule:\\
     - Add an equal distance in opposite directions (e.g., 5 km North and 5 km South), so that it must cancel out the back-and-forth movement.\\
    
  \vspace{1em}

  Your task:\\
  1. Read the base puzzle below.\\
  2. Apply the above transformation and simplification rules.\\
  3. Ensure that the logical movement pattern and final displacement remain the same.\\
  4. Output only the modified (obfuscated) puzzle text — no explanation, no headers or footers.\\

  \vspace{1em}

  Puzzle:\\
  """ \{puzzle\} """\\

\end{llmprompt}

\clearpage

\end{document}